%% file: main.tex
\documentclass[10pt,twocolumn,letterpaper]{article}

\usepackage{iccv}              


\definecolor{iccvblue}{rgb}{0.21,0.49,0.74}
\usepackage[pagebackref,breaklinks,colorlinks,allcolors=iccvblue]{hyperref}
\usepackage{algorithm}
\usepackage{algorithmic}
\usepackage{graphicx}
\usepackage{booktabs}
\usepackage{xcolor}
\usepackage{multirow}
\usepackage{amsfonts}
\usepackage{amsmath}
\usepackage{colortbl}
\usepackage{makecell}
\usepackage[accsupp]{axessibility}  

\def\paperID{*****} 
\def\confName{ICCV}
\def\confYear{2025}

\title{Towards Adversarial Robustness via Debiased High-Confidence Logit Alignment}

\author{
\textbf{Kejia Zhang}$^{1}$, \textbf{Juanjuan Weng}$^{2}\thanks{Corresponding author}$, \textbf{Shaozi Li}$^{1}$, \textbf{Zhiming Luo}$^{1,3}$ \\
\small $^{1}$Department of Artificial Intelligence, Xiamen University, Xiamen, China\\
\small $^{2}$College of Information Science and Technology, Jinan University, Guangzhou, China \\
\small $^{3}$Key Laboratory of Multimedia Trusted Perception and Efficient Computing, Ministry of Education of China, Xiamen University \\
\tt\small kejiaz171@gmail.com, jjweng@jnu.edu.cn, \{szlig,zhiming.luo\}@xmu.edu.cn
}

\begin{document}
\maketitle
\input{sec/0_abstract}    
\input{sec/1_intro}
\input{sec/2_related}
\input{sec/3_method}
\input{sec/4_ex}
\input{sec/5_conclusion}

\section*{Acknowledgment}
This work is supported by the National Natural Science Foundation of China under Grant No.~62276221,~62376232; Key Laboratory of Equipment Data Security and
Guarantee Technology, Ministry of Education under Grant No.~2024020200.

{
    \small
    \bibliographystyle{ieeenat_fullname}
    \bibliography{main}
}

\newpage

\input{sec/appendix}

\end{document}

%% file: sec/0_abstract.tex
\begin{abstract}
Despite the remarkable progress of deep neural networks (DNNs) in various visual tasks, their vulnerability to adversarial examples raises significant security concerns. Recent adversarial training methods leverage inverse adversarial attacks to generate high-confidence examples, aiming to align adversarial distributions with high-confidence class regions. However, our investigation reveals that under inverse adversarial attacks, high-confidence outputs are influenced by biased feature activations, causing models to rely on background features that lack a causal relationship with the labels. This spurious correlation bias leads to overfitting irrelevant background features during adversarial training, thereby degrading the model's robust performance and generalization capabilities. To address this issue, we propose Debiased High-Confidence Adversarial Training (DHAT), a novel approach that aligns adversarial logits with debiased high-confidence logits and restores proper attention by enhancing foreground logit orthogonality. 
Extensive experiments demonstrate that DHAT achieves state-of-the-art robustness on both CIFAR and ImageNet-1K benchmarks, while significantly improving generalization by mitigating the feature bias inherent in inverse adversarial training approaches. Code is available at \url{https://github.com/KejiaZhang-Robust/DHAT}.
\end{abstract}

%% file: sec/1_intro.tex
\section{Introduction}
Deep neural networks (DNNs) have demonstrated exceptional performance across a range of visual tasks~\cite{zeng2024controllable, li2024visual, shao2025holitom, gao2024avsegformer}. 
However, these networks remain susceptible to adversarial attacks, which present significant security challenges~\cite{szegedy2013intriguing, fawzi2018adversarial}.
Adversarial attacks introduce subtle perturbations that are imperceptible to humans but can significantly disrupt the inference process of DNNs, leading to incorrect predictions. 
Adversarial training (AT) is widely recognized as one of the most effective methods for defending against these attacks by incorporating adversarial examples into the training process~\cite{FGSM, shafahi2019adversarial}.

\begin{figure}
    \centering
    \includegraphics[width=\linewidth]{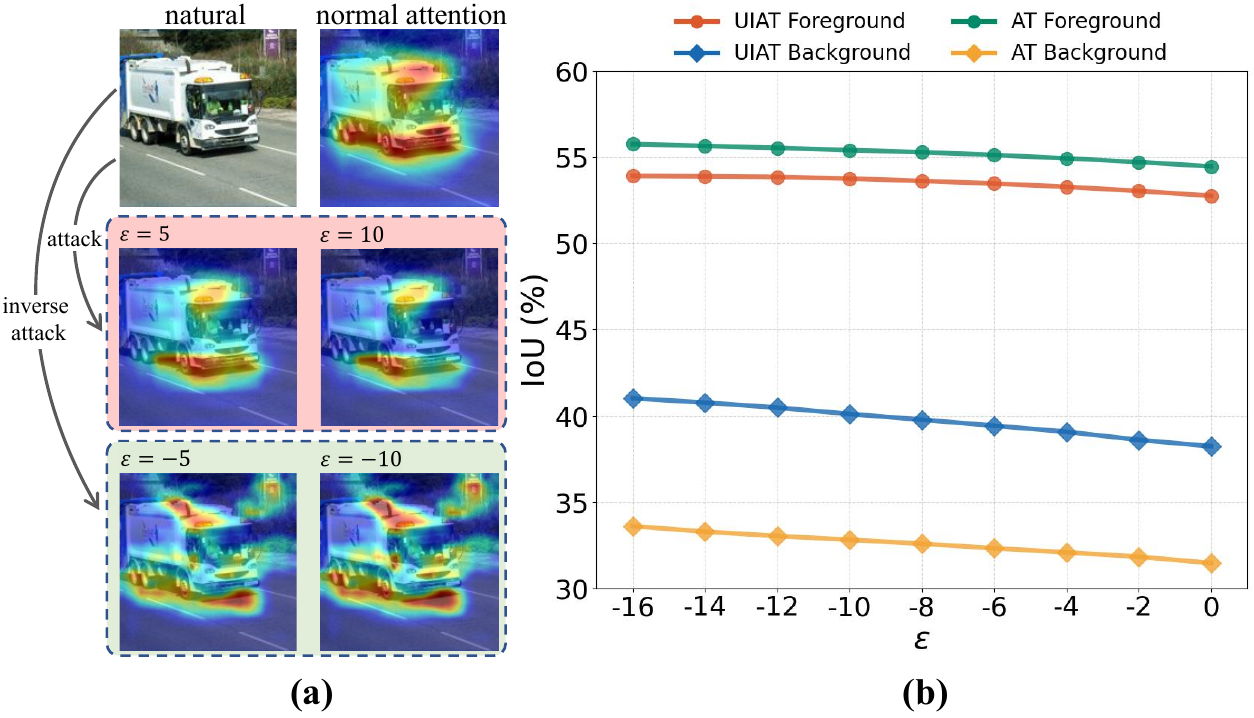}
    \caption{
    Inverse adversarial attack exploits spurious correlations via attention misallocation to non-causal background features. The negative $\epsilon$ denotes inverse adversarial perturbation strength. 
    (a) Grad-CAM~\cite{selvaraju2017grad} visualizations show that adversarial attacks disrupt attention and cause errors, while inverse adversarial attacks amplify activation and improve accuracy but expose a critical flaw: attention shifts from foreground to irrelevant background regions. (b) Quantitative comparison of IoU(\%) between attention regions and the true foreground and background regions using UIAT~\cite{UIAT} and AT. Increasing inverse attack strength $-\epsilon$ demonstrates UIAT's disproportionate focus on non-causal background features, exposing reliance on spurious correlations~\cite{asgari2022masktune,seo2022information,ming2022impact}.
    }
    \label{fig:motivation}
    \vspace{-1em}
\end{figure}

Recently, numerous advanced AT-based techniques have been proposed to address the vulnerabilities of DNNs~\cite{li2023wat, subramanian2024spatial, huang2023fast, li2024asam}. 
Prominent methods in this domain focus on aligning the distributions of adversarial examples to their true classes, thereby enhancing classification accuracy under adversarial attacks. 
For example, MART~\cite{MART} and TRADES~\cite{TRADES} aim to align the logits of adversarial examples with their corresponding natural examples, promoting consistency in predictions. 
Expanding on this, 
UIAT~\cite{UIAT} and ACR~\cite{cho2023anti} extend this approach by generating inverse adversarial examples~\cite{salman2021unadversarial} that exhibit higher confidence compared to their natural counterparts. 
This strategy helps alleviate the negative impact of misclassification of natural examples in alignment and directs adversarial examples toward regions with correct high-confidence classifications.

Upon investigation, we uncover a \textbf{remarkable yet concerning phenomenon}: inverse adversarial examples produce high-confidence outputs by triggering biased feature activation. To illuminate this effect, we visualize attention maps from a conventional adversarial training (AT) model under both adversarial and inverse adversarial attacks (see \Cref{fig:motivation}~(a)). 
Adversarial attacks disrupt normal attention patterns, causing incorrect predictions. In contrast, inverse adversarial attacks amplify the model's attention on features, with stronger attacks even improving prediction accuracy~\cite{UIAT,wen2021defending}. 
However, this apparent enhancement masks a \textbf{critical flaw}: the model's attention systematically shifts from discriminative foreground features to irrelevant background regions—akin to identifying a sheep by detecting grass rather than the animal itself. This reliance on spurious correlations undermines the model's causal reasoning capabilities and compromises its robustness~\cite{izmailov2022feature,crabbe2023evaluating,ahmad2024causal}.

To delve deeper into this issue, we conduct a \textbf{statistical analysis} using the Intersection over Union (IoU) metric to measure the alignment between attention maps and foreground/background regions on a subset of ImageNet~\cite{Howard_Imagenette_2019}. Leveraging the Class Activation Mapping (CAM)-based technique inspired by~\cite{dong2020robust}, we define foreground as regions with high-class activation (\emph{i.e.,} areas critical for predictions) and background as regions with low activation. Comparing the UIAT model~\cite{UIAT} (trained with inverse adversarial examples) to a standard AT model reveals a \textbf{concerning trend}. 
As evidenced in \Cref{fig:motivation}~(b), while increasing the magnitude $\epsilon$ of inverse adversarial attacks enhances overall feature attention, the UIAT model fails to improve IoU for foreground features compared to the AT model. Instead, it demonstrates a marked increase in IoU for background features—a clear indicator of attention misallocation.

This leads to a \textbf{pivotal insight}: the inverse adversarial training disproportionately skews attention toward background features, resulting in biased feature activation. 
This shift engenders a \textbf{fundamental vulnerability}: a spurious correlation bias where the model becomes overly dependent on contextual background cues rather than intrinsic object features. Much like a classifier that associates beaches with birds rather than wings and beaks, this misplaced reliance significantly degrades robustness and generalization performance~\cite{Bias_corre_2023, zhangfeature}, as conclusively demonstrated in~\Cref{st:revrese}. 

Building upon the above findings, we introduce a method named Debiased High-Confidence Adversarial Training (DHAT). 
DHAT implements two key techniques: Debiased High-Confidence Logit Regularization (DHLR) and Foreground Logit Orthogonal Enhancement (FLOE). 
DHLR quantifies the bias towards background feature activation and debiase the spurious correlations by recalibrating the biased high-confidence logits derived from inverse adversarial examples. 
This regularization aligns the logits of adversarial examples with debiased high-confidence logits to improve adversarial robustness and mitigate spurious correlation bias. 
FLOE further refines this process by reducing the correlation between high-confidence logits and background features in affine space, which helps restore the model's attention to its normal state. 
Extensive experiments demonstrate that DHAT outperforms state-of-the-art techniques across various datasets, providing superior adversarial robustness and generalization.
Additionally, our DHAT can be seamlessly integrated with existing advanced adversarial training methods.

Below, we outline our principal contributions: 
\begin{itemize}
    \item 
    We identify that the adversarial training model shows a biased feature activation under inverse adversarial attacks. Training with inverse adversarial examples causes the model's attention shifting towards background features, thus lead to spurious correlation bias.
    \item 
    We introduce a novel method, Debiased High-Confidence Adversarial Training (DHAT). This method aligns the logits of adversarial examples with debiased high-confidence logits corresponding to their classes, effectively mitigating spurious correlation bias.
    \item 
    Extensive experiments validate the superior robustness and generalization of our method against adversarial attacks. Additionally, our approach integrates smoothly with existing advanced adversarial training strategies.
\end{itemize}

%% file: sec/2_related.tex
\section{Related Work}
\subsection{Adversarial Attack}
Adversarial attacks introduce subtle perturbations $\delta$ into natural examples $x$, causing DNNs to produce incorrect predictions rather than the true class $y$.
These attacks are typically formulated as an optimization problem:
\begin{equation}
    \hat{x} = \underset{||\delta||_p \le \epsilon}{\max}(\mathcal{L}_{attack}(\theta;x+\delta,y)),
    \label{attack_max}
\end{equation}
where $\hat{x}$ represents the adversarial example, $\mathcal{L}_{attack}$ is the adversarial attack loss function, $\theta$ denotes the model parameters, and $\delta$ is constrained to the $p$-norm ball of radius $\epsilon$.

Numerous studies have investigated the vulnerability of DNNs and have proposed various adversarial attack methods. 
\citet{FGSM} introduced the Fast Gradient Sign Method (FGSM), which compute the  adversarial perturbation in the direction of the gradient: 
\begin{equation}
    \delta = \eta \cdot \text{sign}(\nabla_x \mathcal{L}_{attack}(\theta; x, y)),
\end{equation}
where $\eta$ denotes the attack step size. Building on this, \citet{PGD_Attack} proposed Projected Gradient Descent (PGD), which iteratively updates the gradient of the input data to generate more potent adversarial examples, thereby increasing the success rate of attacks: 
\begin{equation}
    \hat{x}^{t+1} = \Pi_{\mathbb{B}(\epsilon)} \{ \hat{x}^t + \eta \cdot \text{sign}(\nabla_{\hat{x}^t} \mathcal{L}_{attack}(\theta; \hat{x}^t, y)) \},
\end{equation}
where $\Pi_{\mathbb{B}(\epsilon)}$ projects the perturbation onto the $p$-norm ball of radius $\epsilon$.
Furthermore, \citet{CW} developed the Carlini-Wagner (C\&W) attack, an optimization-based method capable of producing more imperceptible adversarial examples. 
\citet{AA} introduced AutoAttack, an ensemble attack strategy that is user-independent and parameter-free.

\subsection{Adversarial Defense}
Adversarial training (AT) is a widely used strategy for improving model robustness. The core principle of AT is to include adversarial examples in the training process, thereby aligning the model’s outputs towards the true label distribution~\cite{xie2020adversarial,fowl2021adversarial}. 
Mathematically, conventional AT methods are formulated as a min-max optimization problem:
\begin{equation}
    \underset{\theta}{\min}\mathbb{E}_{(x,y)\sim \mathcal{D}}\underset{||\delta||_p \le \epsilon}{\max}\mathcal{L}_{AT}(\theta;\mathcal{G}_{\theta}(x,\delta),y),
\end{equation}
where $(x,y)$ denotes a clean image-label pair sampled from the data distribution $\mathcal{D}$, $\delta$ is constrained to the maximum $p$-norm magnitude $\epsilon$, $\mathcal{L}_{AT}$ represents the training strategy function, and $\mathcal{G}_{\theta}(x,\delta)$ denotes the attack generation procedure incorporating the perturbation $\delta$. The outer minimization optimizes the model parameters $\theta$ to minimize the expected loss over adversarial examples, while the inner maximization identifies the perturbations that maximize this loss, thereby enhancing the model’s robustness.

Recent advancements in AT have introduced several techniques to further improve adversarial robustness. \citet{AWP} introduced Adversarial Weight Perturbation (AWP), which enhances robustness by flattening the loss landscape with respect to input variations. 
\citet{FSR} proposed Feature Separation and Recalibration (FSR) to capture useful logits from non-robust features. \citet{CFA} developed Class-wise Calibrated Fair Adversarial Training (CFA), which adaptively configures training for each class to achieve category-level robust fairness.
\citet{SGLR} proposed SGLR, which uses self-distillation to refine the soft-label distribution logits, thereby calibrating the adversarial training. 

Methods similar to our approach include MART~\cite{MART} and UIAT~\cite{UIAT}. These methods use Kullback–Leibler (KL) divergence to align the logits of adversarial examples with high-confidence logits. 
MART encourages sharing a similar logit distribution between adversarial examples and their corresponding natural examples. 
UIAT generates inverse adversarial examples~\cite{salman2021unadversarial} that exhibit higher confidence compared to natural examples, thereby enhancing the alignment of adversarial examples towards the high-confidence regions of the decision surface. 

%% file: sec/3_method.tex
\section{Methodology}
In this section, we introduce the basic notation and present our approach to mitigate adversarial bias and improve robustness. Our method recalibrates high-confidence logits and reduces reliance on spurious background features. The optimization strategy is outlined, with pseudocode provided in Section 7 of the supplementary material.

\subsection{Notation}
We denote the training dataset as $(X, Y) = \{(x_i, y_i) |~ i \in (1, 2, \dots, N)\}$, where $x$ and $y$ denote the natural input examples and their corresponding labels, respectively. $N$ represents the total number of examples in the dataset. 
Let $\theta$ denote the parameters of a convolutional neural network (CNN) for image classification. The network's output logits are denoted as $f_\theta(\cdot)$, and the final predicted label is given by $\arg\max f_\theta(\cdot)$. 
Inverse adversarial examples $\check{x}$ can be regarded as examples for which the model correctly predicts the label with high-confidence. These examples are generated by applying image perturbations aimed at minimizing the attack loss~\cite{salman2021unadversarial}:
\begin{equation}
    \check{x} = \underset{||\delta||_p \le \epsilon}{\min}(\mathcal{L}_{\text{Inv}}(\theta;x-\delta,y)),
    \label{inv_attack}
\end{equation}
where $\mathcal{L}_{\text{Inv}}$ denotes the loss function used to generate inverse adversarial attack. The logits of the inverse adversarial example $\check{x}$ and the adversarial example $\hat{x}$ are computed as $\check{z} = f_\theta(\check{x})$ and $\hat{z} = f_\theta(\hat{x})$, respectively.

\begin{figure}[t]
    \begin{center}
        \includegraphics[width=\linewidth]{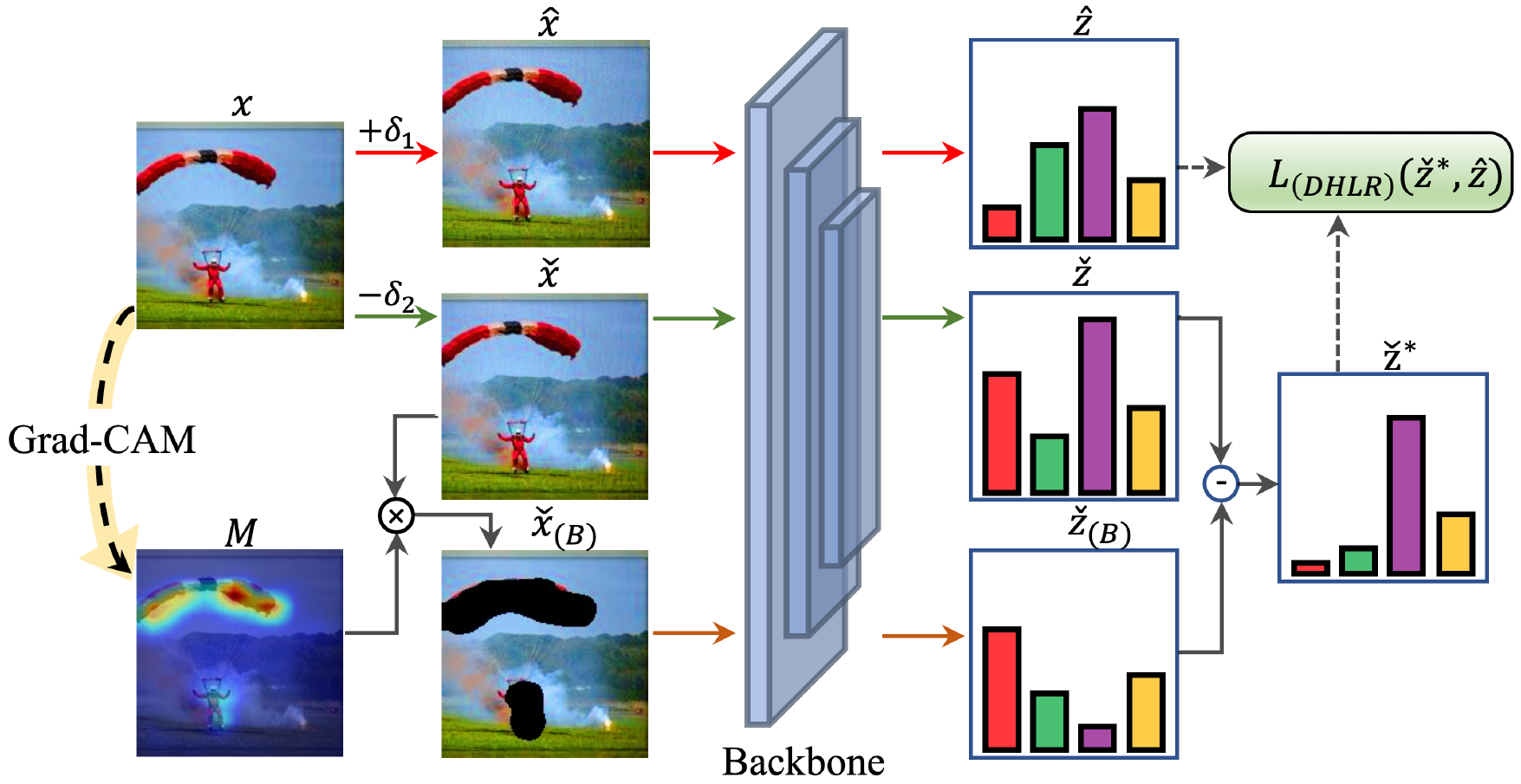}
        \caption{Overview of Debiased High-Confidence Logits Regularization (DHLR). The method first quantifies the degree of biased activation towards background features and recalibrates biased logits, then aligns the logits of adversarial examples with debiased high-confidence logits.}
        \label{Model_overview1}
        \vspace{-1em}
    \end{center}
\end{figure}

\subsection{Debiased High-Confidence Logit Regularization}
To mitigate the spurious correlation bias introduced by adversarial training with inverse adversarial examples, we propose utilizing debiased high-confidence logits for regularization. Specifically, this approach involves separating and quantifying the spurious bias towards background features under inverse adversarial attacks, and then recalibrating the biased high-confidence logits to align with those of adversarial examples, as illustrated in Figure~\ref{Model_overview1}.

First, we quantify the model's biased activation degree towards background features under inverse adversarial attacks. 
Following the approach proposed by \cite{dong2020robust}, we use using Grad-CAM~\cite{grad_cam_paper} or alternative methods such as SAM~\cite{kirillov2023segment} to generate pixel-wise attention maps $M \in [0,1]^{ n\times n}$ for the natural example $x$.
We separate the background feature $\Check{x}_{(B)}$ from the inverse adversarial examples $\Check{x}$ based on the $M$:
\begin{equation}
    [\Check{x}_{(B)}]_{(i,j)} = \mathbb{I}_{(M_{i,j} < \omega)} \cdot \Check{x}_{(i,j)} ,
\end{equation}
where $\mathbb{I}_{(M_{i,j} < \omega)}$ is an indicator function and $\omega$ is a predefined threshold.
Next, we compute the logits for the background feature $\Check{x}_{(B)}$:
\begin{equation}
     \Check{z}_{(B)} = f_{\theta}(\Check{x}_{(B)}).
\end{equation}
These logits can quantify the biased degree introduced by background feature activation during network inference.

To recalibrate the biased high-confidence logits, we refine the logits of inverse adversarial examples by subtracting the background feature logits:
\begin{equation}
    \Check{z}^{*} = \check{z} - \Check{z}_{(B)},
\end{equation}
where $\Check{z}^{*}$ represents the debiased high-confidence logits.

Next, we introduce a regularizer to align the logits of adversarial examples with these debiased high-confidence logits. This regularization encourages the model to focus on foreground features and mitigates the biased activation of background features. Specifically, our Debiased High-Confidence Logit Regularization (DHLR) is defined as: 
\begin{equation}
    \mathcal{L}_{DHLR}(\Check{z}^{*},\Hat{z})= \mathcal{L}_{KL}(\phi(\Check{z}^{*})||\phi(\Hat{z})),
\end{equation}
where $\phi$ denotes the softmax function, $\mathcal{L}_{KL}$ represents the Kullback–Leibler (KL) divergence.

This regularization term $\mathcal{L}_{DHLR}$ promotes consistency between the logits of adversarial examples $\Hat{z}$ and the debiased high-confidence logits $\Check{z}^{*}$. This alignment mitigates spurious correlation bias, thereby improving adversarial robustness and generalization.

\subsection{Foreground Logit Orthogonal Enhancement}
While DHLR effectively recalibrates high-confidence logits to mitigate spurious correlations, it fails to fully restore the model's focus on foreground features when handling inverse adversarial examples $\Check{x}$. This is because DHLR primarily aligns logits with debiased high-confidence targets, without directly addressing the model's persistent bias toward background features. Under inverse adversarial attacks, the model still exhibits strong activation in irrelevant background regions, undermining its ability to prioritize critical foreground cues.

To address this issue, we introduce Foreground Logit Orthogonal Enhancement (FLOE),  a targeted method designed to minimize the correlation between high-confidence logits \(\Check{z}\) and background feature logits \(\Check{z}_{(B)}\). The intuition behind FLOE is straightforward: by reducing the projection of $\Check{z}$ onto $\Check{z}_{(B)}$ in the affine space, we make $\Check{z}$ less explainable by $\Check{z}_{(B)}$,  thereby encouraging the model to rely more on foreground features. This projection-based approach ensures that the model’s predictions are driven by meaningful foreground information rather than spurious background correlations.
The optimization objective is formulated as:
\begin{equation}
    \mathcal{L}_{FLOE}(\Check{z},\Check{z}_{(B)}) = -|\Check{z} - \frac{\Check{z}\cdot \Check{z}_{(B)}}{|\Check{z}_{(B)}|^2} \cdot \Check{z}_{(B)} |_p ,
\end{equation}
where $p$ indicates the norm exponent. 
The term $\mathcal{L}_{FLOE}$ is designed to alleviate biased feature activation under inverse adversarial attacks. 

\subsection{Optimization Strategy}
Finally, we integrate the DHLR and FLOE techniques into a unified training framework Debiased High-Confidence Adversarial Training (DHAT). The overall optimization objective for DHAT is formulated as:
\begin{equation}
    \begin{aligned}
        \mathcal{L}_{DHAT} =& \mathcal{L}_{AT}(\hat{z},y) + \lambda_1 \cdot \mathcal{L}_{DHLR}(\Check{z}^{*},\Hat{z}) \\ &+ \lambda_2 \cdot \mathcal{L}_{FLOE}(\Check{z},\Check{z}_{(B)}), 
    \end{aligned}
\end{equation}
where $\mathcal{L}_{AT}(\hat{z}, y)$ represents the standard adversarial training loss, and $\lambda_1$ and $\lambda_2$ are hyperparameters that control the relative contributions of the DHLR and FLOE terms, respectively. Besides, our DHAT can be seamlessly integrated with various advanced adversarial training strategies represented by $\mathcal{L}_{AT}$ to improve the robustness.

%% file: sec/4_ex.tex
\section{Experiments}

\input{Table/Com_SOTA}

\subsection{Experimental Settings}
\subsubsection{Datasets and models.} 
We conduct experiments on the CIFAR-10, CIFAR-100~\cite{CIFAR}, and ImageNet-1K~\cite{russakovsky2015imagenet} datasets. The architecture for our experiments include the mainstream models WRN28-10~\cite{WRN_NET}, ResNet-18~\cite{ResNet}, ResNet-50~\cite{ResNet}, Inception-V3~\cite{Inception_NET}, and VGG16~\cite{VGG}. Detailed information is provided in the supplementary material.

\subsubsection{Training setup}
For CIFAR datasets, each model is trained for 100 epochs, with the learning rate reduced by a factor of 0.1 at the 80th and 90th epochs. 
For adversarial example generation, we use a fixed 10-iteration attack. The maximum $\ell_\infty$-norm of the adversarial perturbation is set to $\epsilon = 8/255$, while the inverse adversarial perturbation is set to $\epsilon = 4/255$. Both perturbations use a fixed step size of $\alpha = 2/255$. 
For the ImageNet dataset, we adopt the approach in \cite{singh2023revisiting} with 2-iteration PGD over 50 epochs, and set $\epsilon = 4/255$ and $\alpha = 1/255$ for adversarial perturbations.
The hyperparameters $\lambda_1$ and $\lambda_2$ are set to 1.0 in all data sets. Comprehensive training details are available in the supplementary material.

\subsubsection{Evaluation setup.}
For robustness performance evaluation, we employ PGD~\cite{PGD_Attack}, C\&W~\cite{CW}, and AutoAttack (AA)~\cite{AA} attacks under the $\ell_\infty$ norm. AutoAttack comprises APGD-DLR~\cite{AA}, APGD-CE~\cite{AA}, FAB~\cite{FAB}, and Square~\cite{Square}. ``Clean'' denotes natural examples without adversarial perturbations. 
Additionally, we introduce the concept of the robust generalization gap, referred to as the ``Robust Gap'', which measures the difference in robustness performance between the training and test sets as a direct indicator of the generalization ability~\cite{Robust_Gap_Paper}. This metric quantifies the strength of spurious correlations, where larger gaps indicate greater reliance on features that lack causal relationships with target labels.

\input{Table/Diff_net}

\subsection{Experimental Result}
In this part, we conducted comprehensive evaluations of our proposed method across various visual datasets and compared it with mainstream adversarial training (AT) methods. Additionally, DHAT can be integrated with advanced AT strategies. The results are shown in Table~\ref{st:SOTA_Comparison}.
\subsubsection{Robust performance.}
DHAT consistently demonstrates superior robustness across all datasets. For example, on CIFAR-10, DHAT outperforms the second-best method by 1.93\% and 1.16\% under PGD-10 and C\&W attacks, respectively. DHAT also shows the best performance on more complex datasets such as CIFAR-100 and high-resolution Imagenette datasets. Specifically, on ImageNet-1K, DHAT achieves improvements of 1.54\% and 1.52\% over the second-best method under PGD-10 and AA attacks, respectively.

\subsubsection{Combine with other AT methods.}
DHAT can be effectively combined with various AT methods to further enhance adversarial performance. 
Notably, DHAT can be combined with AWP and CFA, denoted as DHAT-AWP and DHAT-CFA, respectively.
However, DHAT is not suitable for combining with methods that rely on spurious correlation to improve performance (\emph{e.g.,} FSR, SLGR).
In particular, DHAT-CFA achieves superior performance across all datasets. 
For instance, on Imagenette, DHAT-CFA provides a 2.84\% performance improvement under the AA attack compared to the second-best method. 
Moreover, combining DHAT with advanced AT surpasses the performance of the original baseline methods in terms of both robustness and clean accuracy.

\subsubsection{Robust overfitting.} 
Our method significantly reduces the robust generalization gap across all evaluated datasets, effectively mitigating the robust overfitting phenomenon. For example, on CIFAR-10, DHAT reduces the robust generalization gap by 4.41\% and 6.01\% compared to UIAT and MART, respectively. This reduction highlights DHAT's ability to mitigate spurious correlation bias, thereby improving the model's overall generalization performance. Detailed visualizations of the learning curve are provided in the supplementary material.

\subsection{Generalization Analysis}
In this part, we evaluate the generalizability of our proposed method by comprehensively evaluating its performance across various network architectures and under different magnitudes of adversarial attacks.

\subsubsection{Performance with diverse network architecture.}
To evaluate the adaptability of our method across different network architectures, we applied DHAT on ResNet-50, VGG-16, and Inception-V3. The performance comparisons are reported in Table~\ref{st:different_net}.
The results demonstrate the DHAT's consistently superior robustness and generalization across a variety of model architectures. For example, DHAT-CFA outperforms the second-best method by 1.83\% under AA attacks on ResNet-50. Similarly, on VGG-16, DHAT significantly reduces the robustness generalization gap by 1.96\% compared to the second-best method.

\begin{figure}[t]
    \centering
    \begin{tabular}{@{}c@{\hspace{-0.25mm}}c@{}}
        \includegraphics[width=0.495\linewidth]{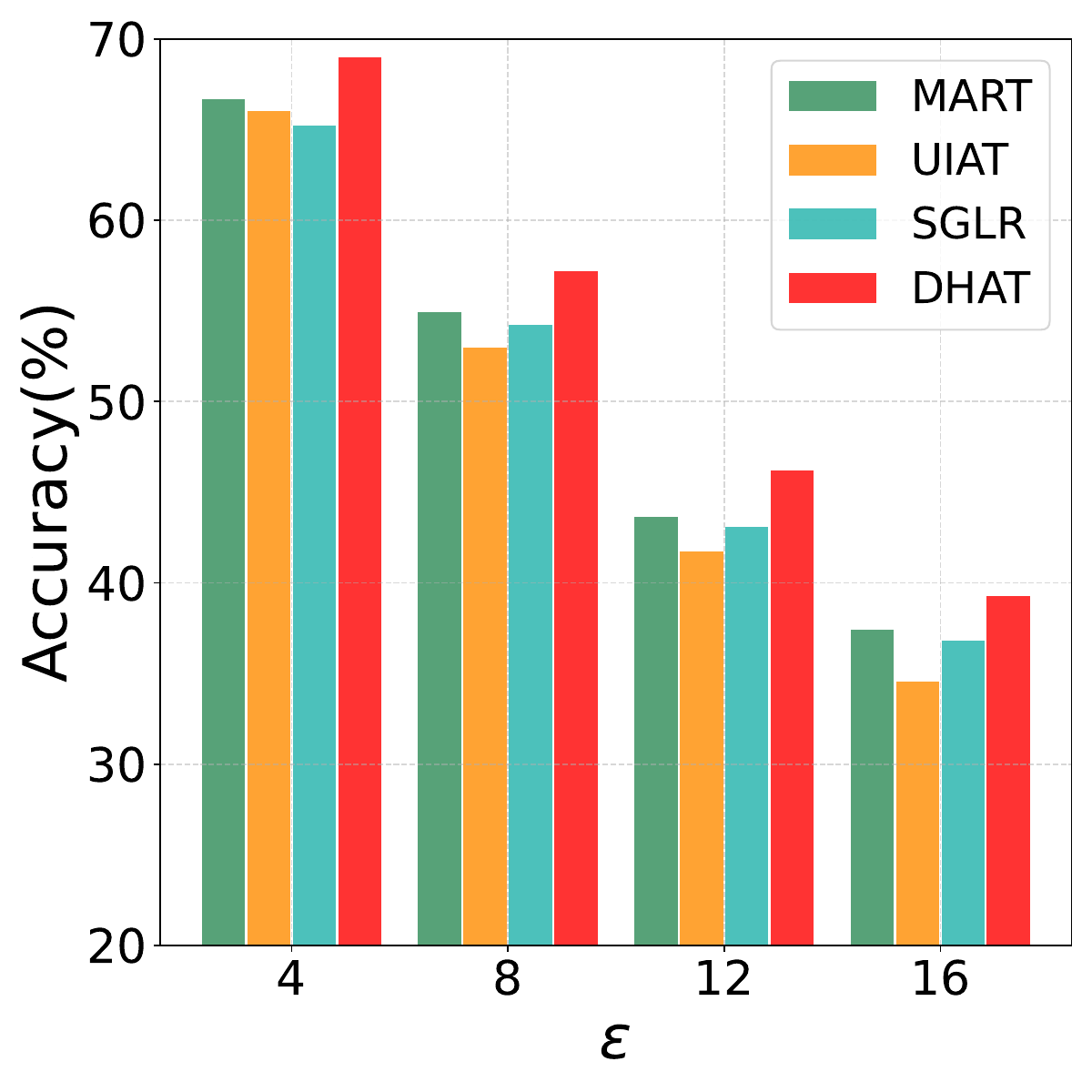}&
        \includegraphics[width=0.495\linewidth]{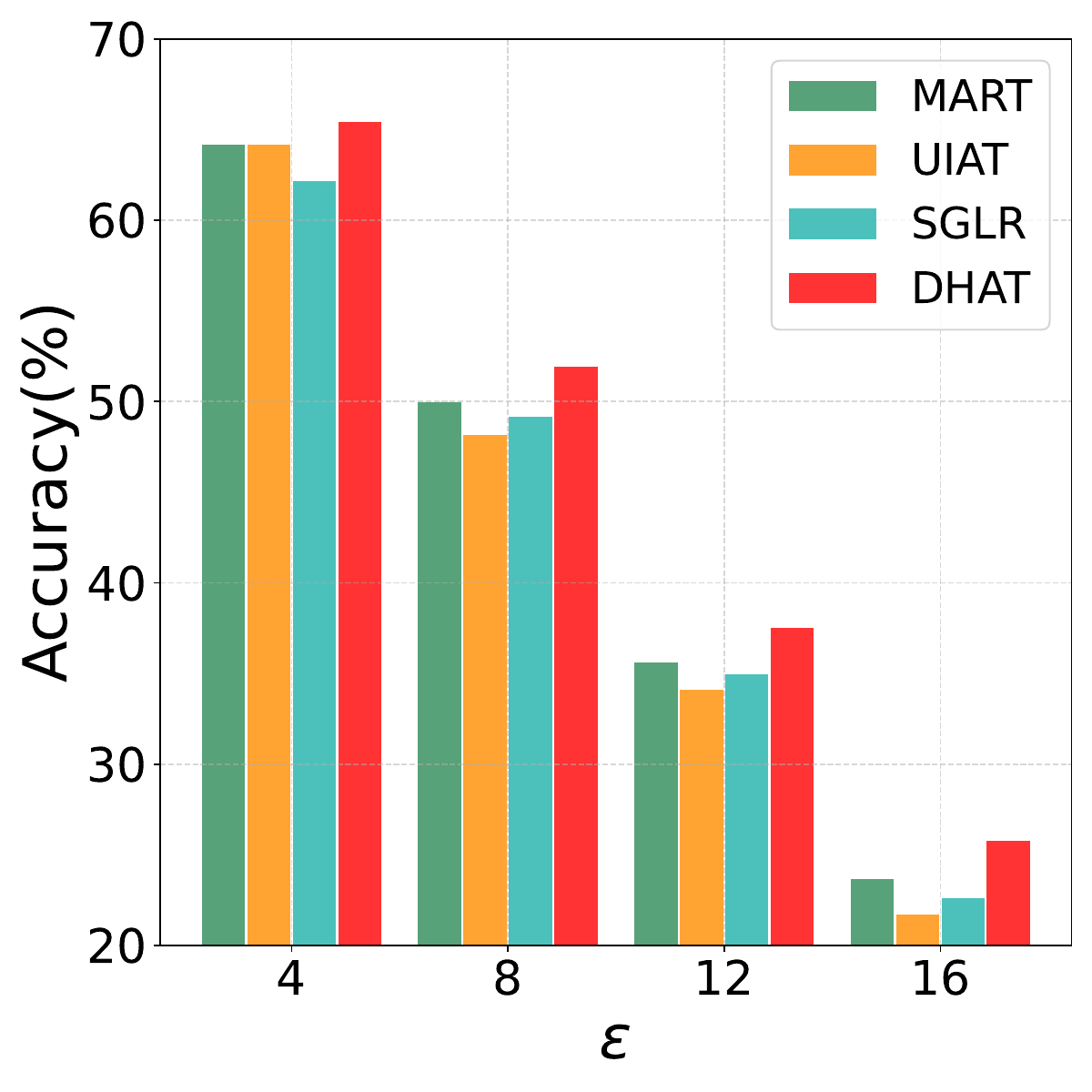}\\
        \parbox{0.495\linewidth}{\centering \footnotesize{\ \ \ \ (a) PGD-10 Attack}} & 
        \parbox{0.495\linewidth}{\centering \footnotesize{\ \ \ \ \ \ (b) C\&W Attack}}
    \end{tabular}
    \caption{Comparisons of robustness~(\%) under varying $\epsilon$ adversarial attacks using ResNet-18 on CIFAR-10.}
    \label{fig:Epsilon_2}
    \vspace{-1em}
\end{figure}

\subsubsection{Different attack magnitude.}
\Cref{fig:Epsilon_2} illustrates the performance of DHAT under varying magnitudes of PGD-10 and C\&W attacks. Our findings reveal that DHAT maintains a significant performance advantage across attack magnitudes. Compared to other methods, DHAT exhibits improved stability and robustness, with minimal performance degradation even under more severe attacks. This consistent performance across varying attack levels underscores the effectiveness of our approach in mitigating the impact of adversarial perturbations. 
Further details are provided in the supplementary material (Section 7).

\input{Table/Computation_cost}

\subsection{Computational Efficiency}
The computational costs of our method and baseline models are provided in terms of per-epoch training time on a single NVIDIA A100 GPU with a batch size of 128 on the CIFAR-10 dataset, as shown in \Cref{tb:computation}. Our method demonstrates significantly improved time efficiency compared to the baseline MART, AWP, FSR, and CFA. While it requires slightly more computational resources than the state-of-the-art methods UIAT and SGLR, this minor increase is justifiable given the performance gains achieved.

\subsection{Ablation Analysis}
In this part, we conduct a comprehensive analysis of the impact of individual components and parameters of our model. 
\subsubsection{Impact of individual components.}
\input{Table/Components}
We evaluate the contributions of the individual components of DHAT, specifically Debiased High-Confidence Logit Regularization (DHLR) and Foreground Logit Orthogonal Enhancement (FLOE), as shown in Table~\ref{tb:componenets}. The results indicate that both DHLR and FLOE significantly enhance model performance compared to the UIAT method. They enhance clean accuracy and adversarial robustness, while also reducing the gap in robustness generalization.
This highlights the effectiveness of DHLR and FLOE in mitigating spurious correlation biases. Notably, the integration of both DHLR and FLOE results in a more substantial overall improvement in model performance.

\begin{figure}[t]
    \centering
    \begin{tabular}{@{}c@{\hspace{-0.25mm}}c@{}}
        \includegraphics[width=0.495\linewidth]{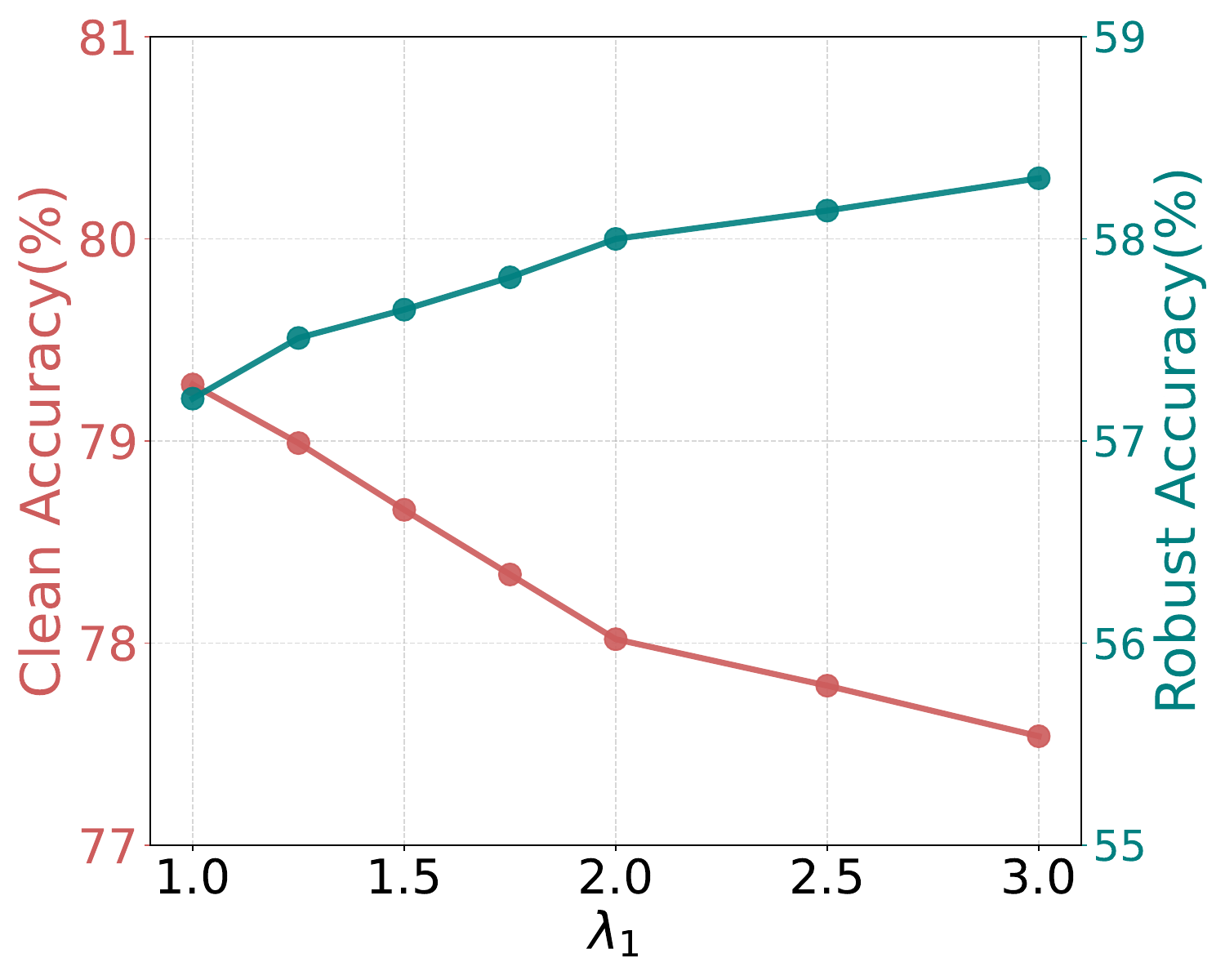}&
        \includegraphics[width=0.495\linewidth]{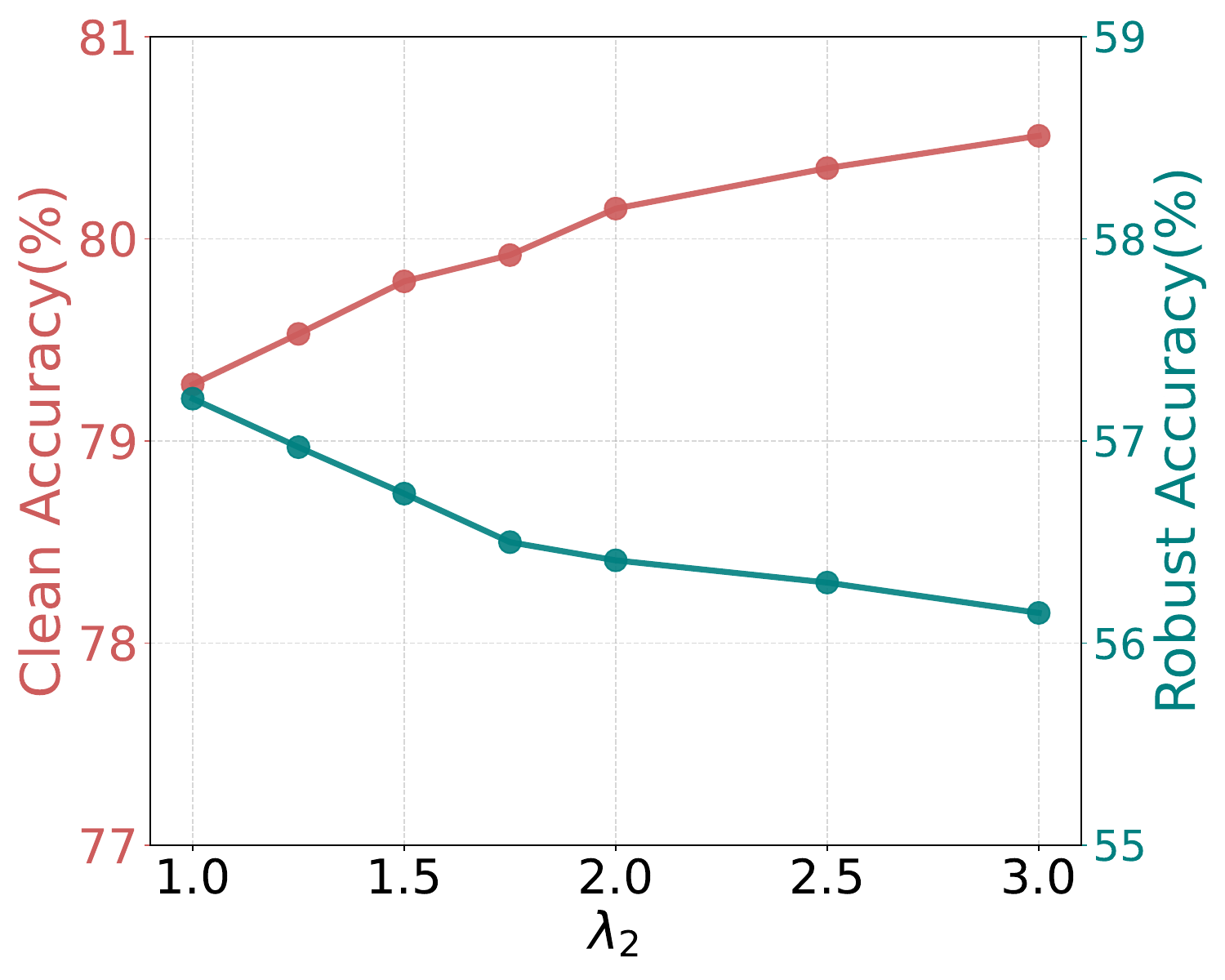}\\
        \parbox{0.495\linewidth}{\centering \footnotesize{\ \  (a) $\mathcal{L}_{DHLR}$}} & 
        \parbox{0.495\linewidth}{\centering \footnotesize{\ \  (b) $\mathcal{L}_{FLOE}$}}
    \end{tabular}
    \vspace{-0.5em}
    \caption{Analysis of hyper-parameters $\lambda_1$ and $\lambda_2$ using ResNet-18 on CIFAR-10. The robust accuracy denotes the accuracy under PGD-10 attack. } 
    \label{fig:Parameter_ana}
    \vspace{-0.0em}
\end{figure}

\subsubsection{Analysis of parameter sensitivity.}
Figure~\ref{fig:Parameter_ana} illustrates the sensitivity of the optimization parameters $\lambda_1$ and $\lambda_2$ in DHAT. Our analysis reveals that these parameters influence the trade-off between adversarial robustness and clean accuracy. 
As $\lambda_1$ increases, adversarial robustness improves, while clean accuracy tends to decrease. This occurs because higher values of $\lambda_1$ increase the regularization strength, which enhances the alignment of adversarial example logits with debiased high-confidence logits. However, this stronger regularization effect may also lead to a reduced emphasis on clean accuracy, as the model focuses more on handling adversarial perturbations. 
Conversely, increasing $\lambda_2$ generally improves clean accuracy but may adversely affect adversarial robustness. Larger values of $\lambda_2$ help the model restore its attention to the normal state, which can result in less effective alignment of adversarial example logits with the debiased high-confidence logits. 

\begin{figure}[t]
    \centering
    \begin{tabular}{@{}c@{\hspace{-0.25mm}}c@{}}
        \includegraphics[width=0.495\linewidth]{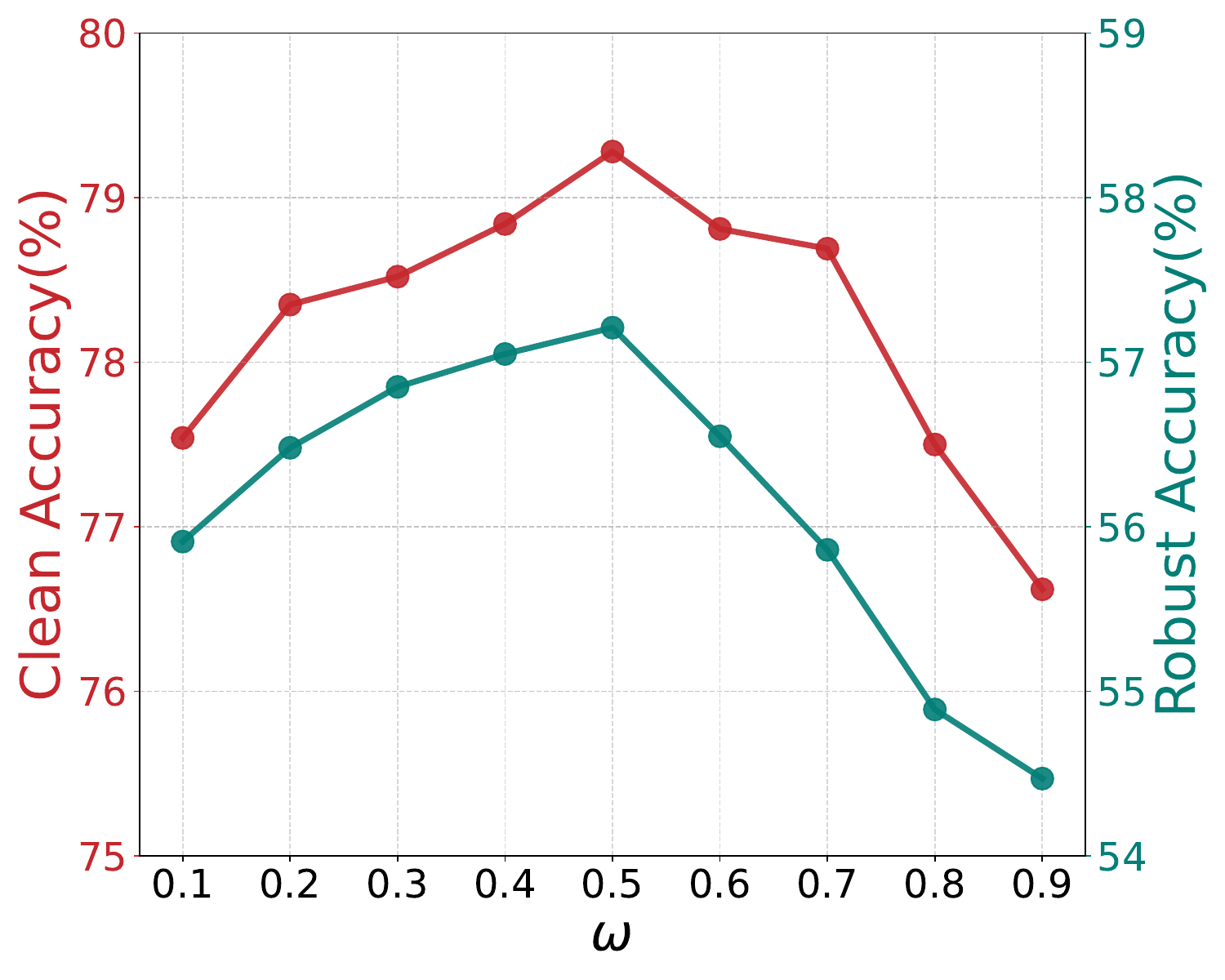}&
        \includegraphics[width=0.495\linewidth]{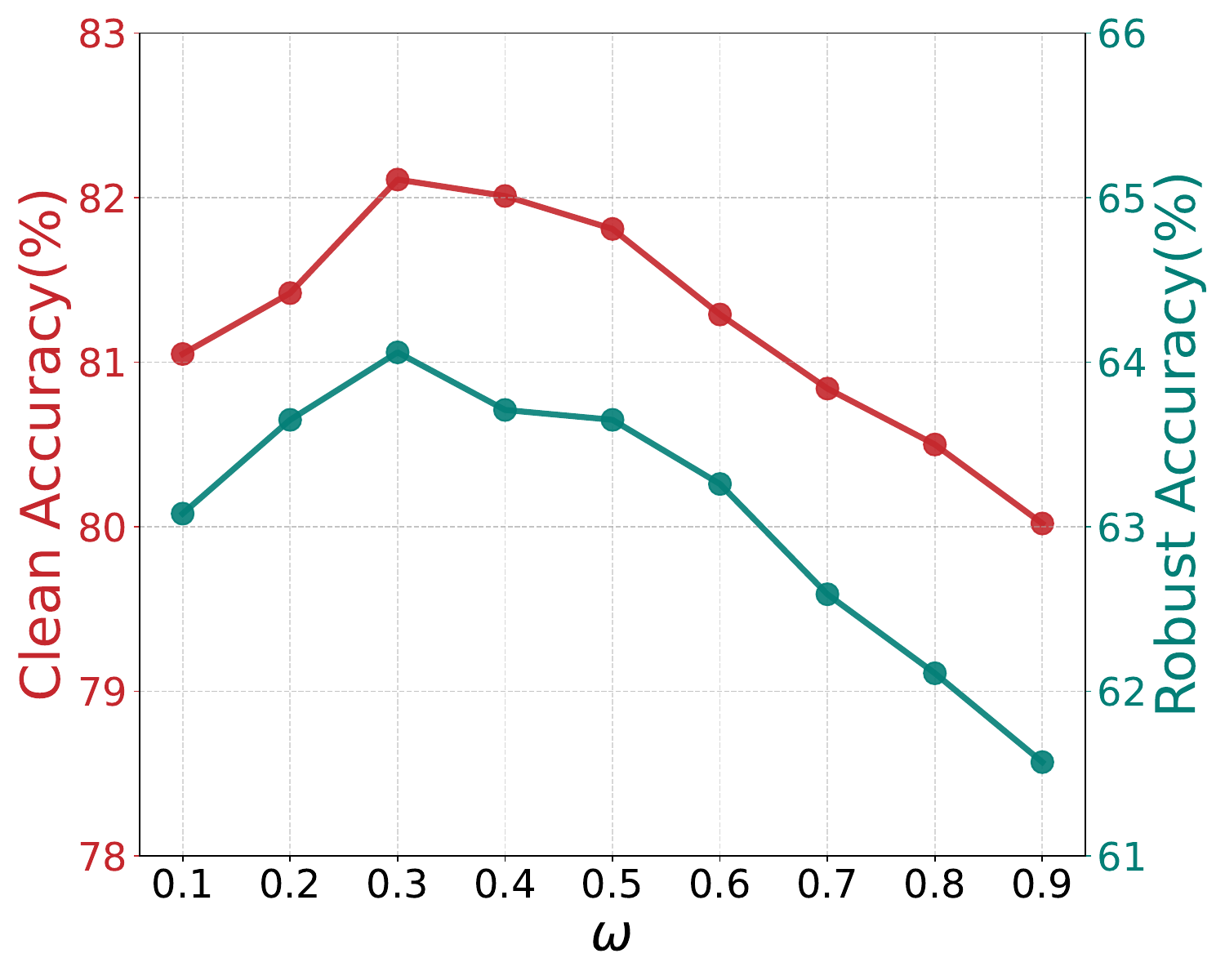}\\
        \parbox{0.495\linewidth}{\centering \footnotesize{(a) CIFAR-10}} & 
        \parbox{0.495\linewidth}{\centering \footnotesize{(b) Imagenette}}
    \end{tabular}
    \vspace{-0.5em}
    \caption{Analysis of hyper-parameters $\omega$ using ResNet-18 on CIFAR-10 and Imagenette.  The robust accuracy denotes the accuracy under PGD-10 attack. }
    \label{fig:threshold_ana}
    \vspace{-0.0em}
\end{figure}

\subsubsection{Analysis of threshold factor for quantifying spurious bias.}
Figure~\ref{fig:threshold_ana} illustrates the impact of threshold values $w$ on datasets with varying resolutions. Our analysis reveals that as the threshold value increases, both clean accuracy and robustness initially improve before sharply declining.
This initial improvement occurs because higher thresholds help in better separating accurate background features. However, further increasing the threshold eventually leads to the extraction of more foreground features, which causes significant performance degradation. For low-resolution datasets, the optimal threshold tends to be higher compared to high-resolution datasets, as high-resolution images more readily separate background features.

\begin{figure}[t]
    \begin{center}
        \includegraphics[width=\linewidth]{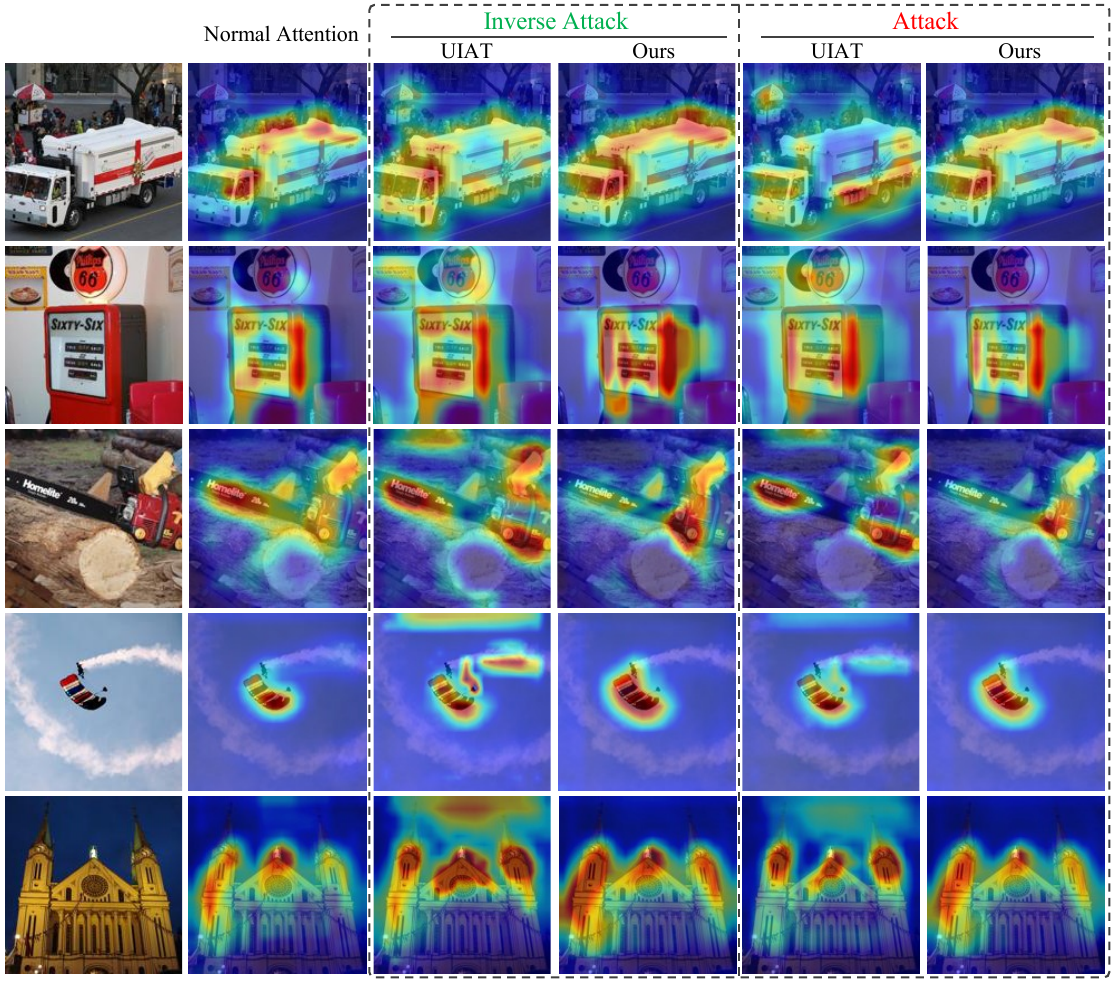}
        \caption{Visualization of feature activation maps using UIAT and DHAT, utilizing Grad-CAM under inverse attack ($\epsilon=-8$) and adversarial attack ($\epsilon=8$). 
        Left-to-right: the original image; normal feature activation maps; feature activation maps under adversarial attacks for UIAT and DHAT; and feature activation maps under inverse adversarial attacks for UIAT and DHAT.
        }
        \label{fig:CAM_comparison}
    \end{center}
    \vspace{-1.5em}
\end{figure}

\subsection{Discussion DHAT Effectiveness}
In this part, we analyze the mechanisms underlying the effectiveness of our proposed method, Debiased High-Confidence Adversarial Training (DHAT), with a focus on its capacity to mitigate spurious correlation bias.

\subsubsection{Feature Activation Map Analysis}
\Cref{fig:CAM_comparison} visualizes feature activation maps of UIAT and DHAT models under adversarial and inverse adversarial attacks. DHAT successfully maintains normal attention patterns, while UIAT exhibits biased activation with attention shifting significantly to background features during inverse attacks. DHAT achieves this robust performance by recalibrating biased high-confidence logits and systematically reducing correlation between logits and background features.


\begin{figure}[t]
    \centering
    \begin{tabular}{@{}c@{\hspace{-0.25mm}}c@{}}
        \includegraphics[width=0.495\linewidth]{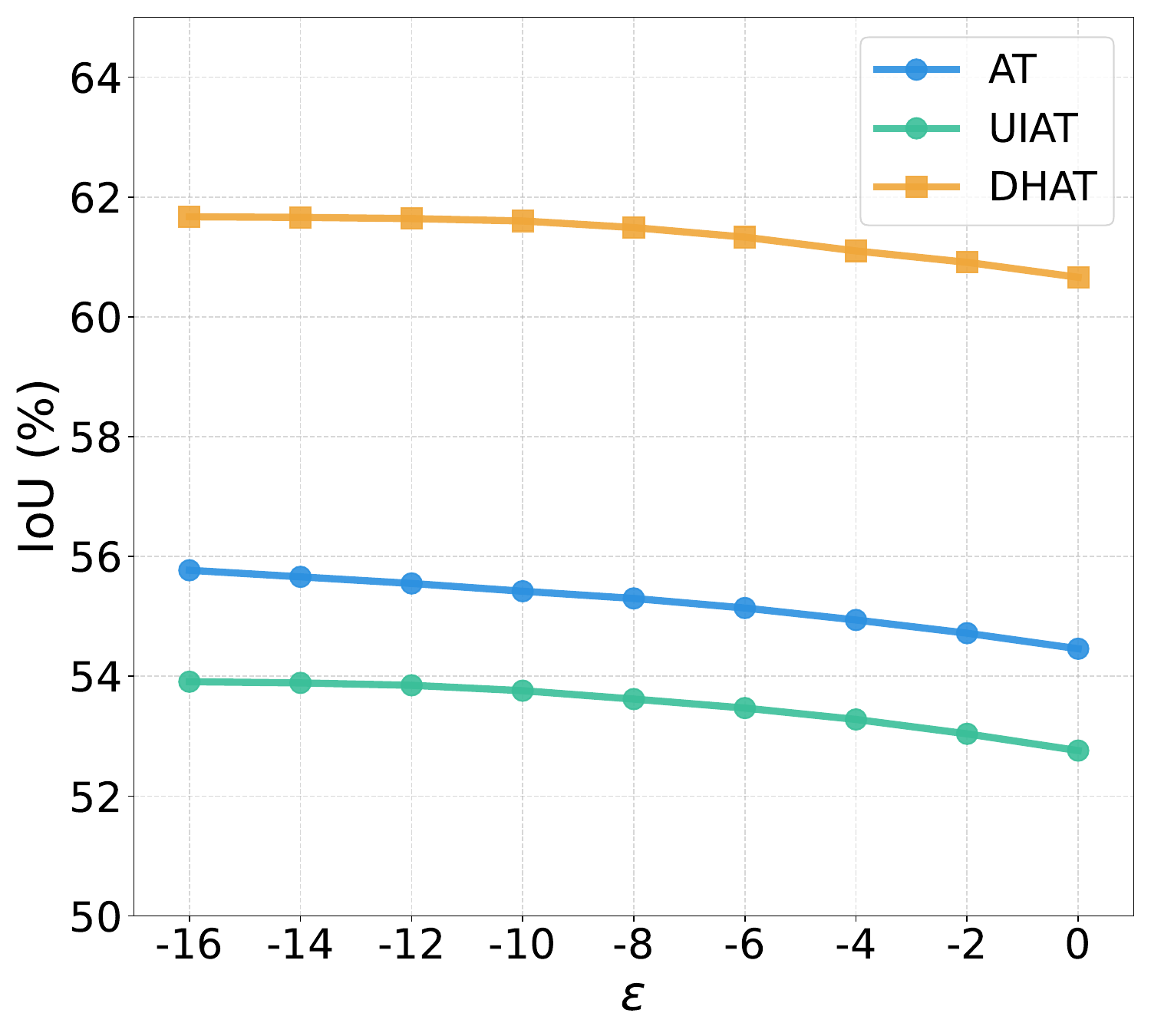}&
        \includegraphics[width=0.495\linewidth]{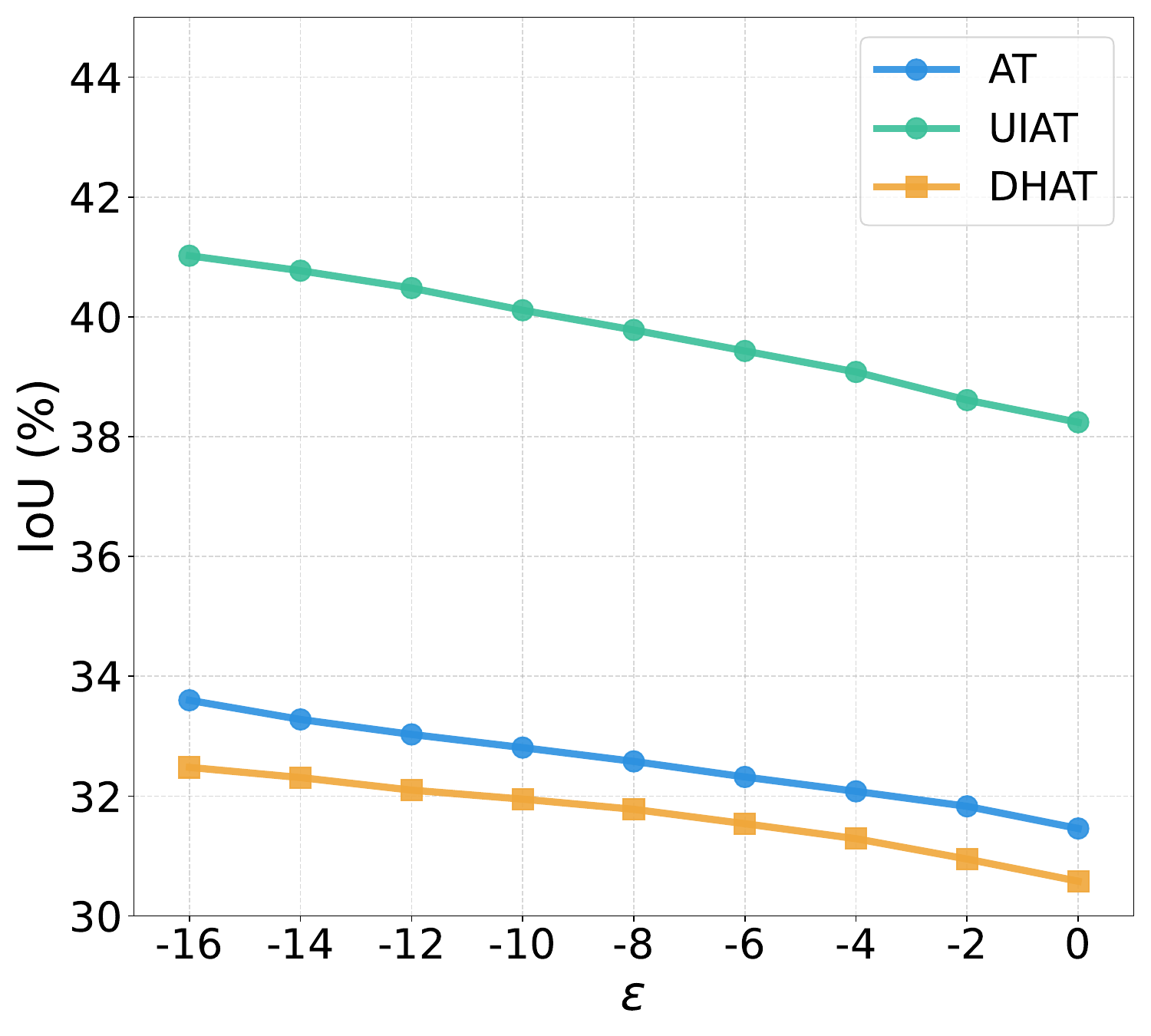}\\
        \parbox{0.495\linewidth}{\centering \footnotesize{ \ (a) IoU with Normal Foreground}} & 
        \parbox{0.495\linewidth}{\centering \footnotesize{ \ \ \  (b) IoU with Normal Background}}
    \end{tabular}
    \vspace{-0.5em}
    \caption{IoU of attention regions with normal foreground and background regions. The comparison is conducted among conventional AT, UIAT, and DHAT on the Imagenette Dataset.}
    \label{fig:ex_motivation}
\end{figure}

\input{Table/UIAT_GAP}

\subsubsection{Quantitative Evaluation of Bias Mitigation}
\Cref{fig:ex_motivation} quantifies DHAT's effectiveness through IoU between attention maps and true foreground/background regions. DHAT achieves higher foreground IoU and lower background IoU than both AT and UIAT, confirming its superior feature attention. \Cref{st:revrese} shows stronger inverse attacks in UIAT worsen overfitting, increasing the robust gap from 9.38 to 14.43. In contrast, DHAT achieves better test robustness (83.95\% PGD-10, 53.10\% AA) with a significantly smaller robust gap (3.51), effectively mitigating overfitting while maintaining strong performance.


%% file: Table/Com_SOTA.tex
\begin{table*}[t]
        \begin{center}
        \caption{Comparison of robustness~(\%) and robust generalization gap~(\%). The \textbf{number} in bold indicates the best performance.}
        \resizebox{0.85\textwidth}{!}
        {
            \begin{tabular}{lccccccc}
            \toprule[1.5pt] 
            \textbf{Architecture: WRN28-10} & \multicolumn{6}{c}{Attack~($\epsilon=4/255$)} &\multicolumn{1}{c}{\multirow{2}{*}{Robust Gap$\downarrow$}}\\
            \cmidrule{2-7}
            \textbf{ImageNet-1K} & Clean$\uparrow$& PGD-10$\uparrow$&PGD-20$\uparrow$&PGD-50$\uparrow$&C\&W$\uparrow$&AA$\uparrow$&\\ 
            \midrule
            MART~\cite{MART}   &62.31&43.56&42.89&42.54&40.62&38.94&13.76\\
            AWP~\cite{AWP}     &64.25&45.13&44.67&44.21&41.96&40.02&12.82\\
            FSR~\cite{FSR}     &64.70&44.31&43.82&43.45&41.23&39.30&14.17\\
            CFA~\cite{CFA}     &65.38&44.87&44.32&43.96&41.58&39.65&15.49\\
            UIAT~\cite{UIAT}   &62.64&45.29&44.85&44.57&42.13&40.18&14.68\\
            SGLR~\cite{SGLR}   &63.46&44.52&44.05&43.78&41.36&39.47&19.72\\
            \rowcolor{gray!20}
            DHAT~(Ours)        &65.90&46.83&46.42&46.11&43.25&41.70&\textbf{9.53}\\
            \rowcolor{gray!20}
            DHAT-AWP~(Ours)    &65.62&47.38&47.04&46.79&43.62&42.07&10.27\\
            \rowcolor{gray!20}
            DHAT-CFA~(Ours)    &\textbf{66.26}&\textbf{48.27}&\textbf{47.89}&\textbf{47.53}&\textbf{44.18}&\textbf{42.45}&11.64\\
            \midrule[1.5pt]
            \textbf{Architecture: WRN28-10} & \multicolumn{6}{c}{Attack~($\epsilon=8/255$)} &\multicolumn{1}{c}{\multirow{2}{*}{Robust Gap$\downarrow$}}\\
            \cmidrule{2-7}
            \textbf{CIFAR-10}&Clean$\uparrow$& PGD-10$\uparrow$&PGD-20$\uparrow$&PGD-50$\uparrow$&C\&W$\uparrow$&AA$\uparrow$&\\ 
            \midrule
            MART~\cite{MART}     &82.99&56.25&55.48&55.45&52.26&50.67&9.52\\
            AWP~\cite{AWP}       &82.67&57.80&57.21&57.07&54.82&51.90&6.90\\
            FSR~\cite{FSR}       &82.92&56.69&55.94&55.51&53.93&51.74&7.42\\
            CFA~\cite{CFA}       &84.43&57.87&56.90&56.64&54.60&51.85&10.36\\
            UIAT~\cite{UIAT}     &82.94&58.66&58.12&58.05&54.11&52.17&7.92\\
            SGLR~\cite{SGLR}     &\textbf{85.76}&57.53&56.91&56.66&54.28&52.07&9.38\\
            \rowcolor{gray!20}
            DHAT~(Ours)          &83.95&60.49&59.95&59.87&55.27&53.10&\textbf{3.51}\\
            \rowcolor{gray!20}
            DHAT-AWP~(Ours)      &83.21&61.61&61.35&61.27&55.71&53.69&4.54\\
            \rowcolor{gray!20}
            DHAT-CFA~(Ours)                 &84.49&\textbf{62.67}&\textbf{62.38}&\textbf{62.22}&\textbf{55.95}&\textbf{54.05}&6.33\\
            \midrule[1.5pt]
            \textbf{Architecture: WRN28-10} & \multicolumn{6}{c}{Attack~($\epsilon=8/255$)} &\multicolumn{1}{c}{\multirow{2}{*}{Robust Gap$\downarrow$}}\\
            \cmidrule{2-7}
            \textbf{CIFAR-100}&Clean$\uparrow$&PGD-10$\uparrow$&PGD-20$\uparrow$&PGD-50$\uparrow$&C\&W$\uparrow$&AA$\uparrow$&\\ 
            \midrule
            MART~\cite{MART}   &54.69&32.06&31.90&31.88&28.77&27.25&9.96\\
            AWP~\cite{AWP}     &57.94&34.01&33.75&33.72&30.74&28.90&7.87\\
            FSR~\cite{FSR}     &57.48&32.93&32.30&32.26&29.16&27.04&7.84\\
            CFA~\cite{CFA}     &60.92&33.10&32.56&32.41&30.49&28.04&10.47\\
            UIAT~\cite{UIAT}   &57.65&34.27&33.91&33.85&30.97&29.03&11.70\\
            SGLR~\cite{SGLR}   &61.02&33.43&32.98&32.82&30.72&28.50&15.67\\
            \rowcolor{gray!20}
            DHAT~(Ours)                &59.14&35.82&35.33&35.02&31.72&30.17&\textbf{4.24}\\
            \rowcolor{gray!20}
            DHAT-AWP~(Ours)               &59.30&36.39&36.15&35.98&32.05&30.38&4.85\\
            \rowcolor{gray!20}
            DHAT-CFA~(Ours)               &\textbf{61.54}&\textbf{37.67}&\textbf{37.15}&\textbf{36.99}&\textbf{32.40}&\textbf{30.93}&5.93\\            
            \bottomrule[1.5pt]
            \end{tabular}
        }
        \label{st:SOTA_Comparison}
        \vspace{-1em}
        \end{center}
\end{table*}

%% file: Table/Diff_net.tex
\begin{table*}[t]
    \begin{center}
    \vspace{-0em}
    \caption{Comparison of robustness~(\%) and robust generalization gap~(\%). The \textbf{number} in bold indicates the best performance.}
    \label{st:different_net}
    \resizebox{0.95\textwidth}{!}
    {
        \begin{tabular}{lccccccccc}
        \toprule[1.5pt] 
        Dataset: CIFAR-10 & \multicolumn{3}{c}{ResNet-50} & \multicolumn{3}{c}{VGG-16} & \multicolumn{3}{c}{Inception-V3}\\
        \cmidrule(lr){2-4} \cmidrule(lr){5-7} \cmidrule(lr){8-10}
        Method& Clean$\uparrow$ & AA$\uparrow$ & Robust Gap$\downarrow$ & Clean$\uparrow$ & AA$\uparrow$ & Robust Gap$\downarrow$ & Clean$\uparrow$ & AA$\uparrow$ & Robust Gap$\downarrow$ \\
        \midrule
        MART~\cite{MART}     &75.05&49.05&4.57&66.72&44.27&3.60&76.05&49.59&5.91\\
        AWP~\cite{AWP}       &75.59&50.80&3.91&67.14&45.97&3.47&76.74&50.46&4.07\\
        FSR~\cite{FSR}       &78.38&50.59&3.97&69.59&44.90&3.30&77.28&50.02&4.73\\
        CFA~\cite{CFA}       &77.88&50.64&4.27&70.45&45.53&4.66&76.70&49.97&4.46\\
        UIAT~\cite{UIAT}     &78.42&51.00&2.99&68.46&45.27&3.57&75.19&51.23&3.70\\
        SGLR~\cite{SGLR}     &80.39&50.55&5.21&\textbf{70.90}&45.04&3.94&\textbf{80.62}&50.73&5.00 \\
        \rowcolor{gray!20}
        DHAT~(Ours)                  &79.94&51.92&\textbf{1.85}&70.01&46.40&\textbf{1.34}&77.85&51.81&\textbf{1.92}\\
        \rowcolor{gray!20}
        DHAT-AWP~(Ours)                 &76.84&52.05&2.34&67.41&47.21&2.18&77.12&52.08&2.88\\
        \rowcolor{gray!20}
        DHAT-CFA~(Ours)                 &\textbf{80.97}&\textbf{52.38}&2.09&70.52&\textbf{47.83}&2.21&78.52&\textbf{52.67}&2.44\\
        \bottomrule[1.5pt]
        \end{tabular}
    }
    \end{center}
    \vspace{-1em}
\end{table*}

%% file: Table/Computation_cost.tex
\begin{table}[t]
    \begin{center}
     \caption{
        Comparison of computational costs (\# A100 GPU hours) using different adversarial training methods on CIFAR-10. 
    }
    \label{tb:computation}
    \resizebox{0.9\columnwidth}{!}
    {
        \begin{tabular}{l c c c}
            \toprule
            \multicolumn{1}{c}{\multirow{2}{*}{Method}} & \multicolumn{3}{c}{Architecture}\\
            \cmidrule{2-4}
             ~ & ResNet-18 & WRN28-10 & Inception-V3 \\
            \midrule
            MART~\cite{MART}  & 0.050 & 0.285 & 0.331 \\
            AWP~\cite{AWP}    & 0.048 & 0.274 & 0.318 \\
            FSR~\cite{FSR}    & 0.051 & 0.291 & 0.337 \\
            CFA~\cite{CFA}    & 0.045 & 0.262 & 0.303 \\
            UIAT~\cite{UIAT}  & 0.040 & 0.245 & 0.283 \\
            SGLR~\cite{SGLR}  & 0.039 & 0.240 & 0.278 \\
            \rowcolor{gray!20}
            DHAT (Ours)       & 0.041 & 0.248 & 0.287 \\
            \bottomrule
        \end{tabular}}
    \end{center}
\end{table}

%% file: Table/Components.tex
\begin{table}[t]
        \begin{center}
        \caption{Comparison of robustness~(\%) and robust generalization gap~(\%). The \textbf{number} in bold indicates the best performance.}
        \resizebox{0.9\linewidth}{!}
        {
            \begin{tabular}{cc|cccc}
            \toprule
            DHLR&FLOE&Clean$\uparrow$&PGD-10$\uparrow$&AA$\uparrow$&Robust Gap$\downarrow$\\ 
            \midrule
            &&78.03&55.03&48.35&3.84\\
            \checkmark&&78.79&56.56&49.47&1.98\\
            &\checkmark&78.50&56.22&49.73&2.06\\
            \rowcolor{gray!20}
            \checkmark&\checkmark&\textbf{79.28}&\textbf{57.21}&\textbf{50.25}&\textbf{0.81}\\
            \bottomrule
            \end{tabular}
        }
        \label{tb:componenets}
        \vspace{-1em}
        \end{center}
\end{table}

%% file: Table/UIAT_GAP.tex
\begin{table}[t]
    \begin{center}
    \caption{Comparison of training with different magnitude reverse adversarial samples (UIAT) and our debiased training DHAT using WRN28-10 on the CIFAR-10.}
    \label{st:revrese}
    \resizebox{\linewidth}{!}
    {
        \begin{tabular}{lccccccccc}
        \toprule[1.5pt] 
        \multicolumn{1}{c}{\multirow{2}{*}{Method}} & \multicolumn{1}{c}{\multirow{2}{*}{$\epsilon$ of $\Hat{z}$}} & \multicolumn{2}{c}{Training Set} & \multicolumn{2}{c}{Test Set} & \multicolumn{1}{c}{\multirow{2}{*}{Robust Gap$\downarrow$}} \\
        \cmidrule(lr){3-4} \cmidrule(lr){5-6}
        && PGD-10$\uparrow$ & AA$\uparrow$ & PGD-10$\uparrow$ & AA$\uparrow$ & \\
        \midrule[0.4pt]
        UIAT&-4 &90.53&61.45&82.94&52.07&9.38\\
        UIAT&-8 &92.57&62.87&82.26&51.33&11.54\\
        UIAT&-12&94.90&65.28&81.72&50.85&14.43\\
        \rowcolor{gray!20}
        DHAT~(ours)&-4 &\textbf{88.73}&\textbf{56.61}&\textbf{83.95}&\textbf{53.10}&\textbf{3.51} \\
        \bottomrule[1.5pt]
        \end{tabular}
   }
    \end{center}
    \vspace{-1em}
\end{table}

%% file: sec/5_conclusion.tex
\section{Conclusion}
In this work, we identify a critical issue in adversarial training: AT model exhibits biased feature activation under inverse adversarial attacks. Specifically, training with inverse adversarial examples causes the model attention to shift from the foreground to background features, resulting in spurious correlation bias.
To tackle this challenge, we propose a novel method called Debiased High-Confidence Adversarial Training (DHAT). 
DHAT incorporates two pivotal techniques: Debiased High-Confidence Logit Regularization~(DHLR) and Foreground Logit Orthogonal Enhancement~(FLOE). 
DHLR quantifies the bias towards background feature activation and recalibrates biased high-confidence logits to align the logits of adversarial examples with these debiased high-confidence logits. 
FLOE further restores the model's attention to its normal state by reducing the correlation between high-confidence logits and background features in affine space.
This work highlights DHAT's effectiveness in enhancing model resilience and presents an adversarial training framework that integrates seamlessly with advanced training strategies.

%% file: sec/appendix.tex
\section*{Supplementary Materials}

This supplementary material offers a detailed examination of the methodologies and results that underpin the experiments in our study. It is designed to provide comprehensive information to validate the findings and ensure reproducibility and transparency.

The content is organized as follows:

\begin{itemize}
    \item \textbf{Section~\ref{sec:Ex}:} A comprehensive and detailed description of the experimental setup and methodology.
    \item \textbf{Section~\ref{sec:algorithm}:} A formal presentation of the proposed algorithm, including its pseudocode representation.
    \item \textbf{Section~\ref{sec:trans_task}:} In-depth evaluation of model performance under various transfer attack scenarios.
    \item \textbf{Section~\ref{sec:generated}:} Detailed results from adversarial training using synthetic data for robustness assessment.
    \item \textbf{Section~\ref{sec:recognition}:} Extensive exploration of techniques for generating foreground-background attention maps.
    \item \textbf{Section~\ref{sec:epsilon}:} Thorough examination of the model's robustness under varying levels of attack intensity.
\end{itemize}

\section{Experimentation Details} \label{sec:Ex}
In this part, we provide a comprehensive overview of the experimental setup. 

\subsection{Training Setup}
\subsubsection{CIFAR-10 and CIFAR-100}
For the CIFAR-10 and CIFAR-100~\cite{CIFAR} datasets, we employ the Stochastic Gradient Descent (SGD) optimizer with a momentum of 0.9 and a weight decay factor of $5 \times 10^{-4}$. The initial learning rate is set to 0.1, and models undergo 100 epochs of training with the learning rate reduced by a factor of 0.1 at the 80th and 90th epochs.

For adversarial training on these datasets, we create adversarial examples through a 10-iteration attack where the maximum $\ell_\infty$-norm of the adversarial perturbation is limited to $\epsilon = 8/255$, using a step size of $\alpha = 2/255$. To ensure reliability and fairness, we align the inverse adversarial example generation process with the UIAT settings~\cite{UIAT}, employing the same loss function and perturbation constrained by $\epsilon = 4/255$.

\subsubsection{ImageNet-1K}
For the ImageNet-1K dataset, we strictly follow the training protocol established in \cite{singh2023revisiting} to ensure fair comparison with existing methods. Specifically, we implement a 2-iteration PGD over 50 epochs, and set $\epsilon = 4/255$ and $\alpha = 1/255$ for adversarial perturbations. This adherence to established benchmarking parameters enables direct and meaningful comparison with state-of-the-art approaches evaluated on this challenging large-scale dataset.

\subsubsection{Common Settings}
The hyperparameters $\lambda_1$ and $\lambda_2$ are consistently set to 1.0 across all datasets. To maintain experimental fairness, all comparative methods undergo identical training strategies with the same number of epochs and optimizer configurations. For Grad-CAM to quantify the spurious correlation bias, we use a pre-trained ResNet-18 model under simple training, introducing minimal additional training cost.

All experiments were conducted on a single NVIDIA Tesla A100. Notably, our method does not employ label smoothing techniques to avoid potential conflicts with efforts to mitigate spurious correlation bias, as label smoothing can blur class boundaries and undermine the precision required for robust decision-making. Additionally, unlike UIAT, our approach does not utilize momentum terms to stabilize the generation process of inverse adversarial examples, instead focusing solely on the intrinsic properties and underlying dynamics of our method. Comprehensive training details are available in the supplementary material.

\subsection{Evaluation Setup}
Our evaluation consists of two main components: robustness performance and robust generalization. To evaluate robustness performance, we use PGD~\cite{PGD_Attack}, C\&W~\cite{CW}, and AutoAttack (AA)~\cite{AA} within the $\ell_\infty$-norm. AutoAttack includes several attack methods such as APGD-DLR~\cite{AA}, APGD-CE~\cite{AA}, FAB~\cite{FAB}, and Square~\cite{Square}. Adversarial attacks are generated using a step size of $\alpha = 2/255$ and a specified maximum $\ell_\infty$-norm. 
Note that ``Clean'' indicates natural examples unaffected by adversarial perturbations.

To evaluate robust generalization, we introduce the concept of the robust generalization gap~\cite{Robust_Gap_Paper} denoted as the ``Robust Gap'' quantifying the difference in robust performance between training and test sets under adversarial attacks. A smaller robust gap indicates improved robust generalization, reflecting reduced vulnerability to robust overfitting in the adversarial training model.

\section{Algorithm Details}
\label{sec:algorithm}
\begin{algorithm}[t]
\caption{Debiased High-Confidence Adversarial Training (DHAT)}
\label{alg:dhat}
\begin{algorithmic}[1]
\REQUIRE Training dataset $(X, Y) = \{(x_i, y_i)\}_{i=1}^N$; Network parameters $\theta$; Perturbation budget $\epsilon$; Hyperparameters $\lambda_1, \lambda_2$; Attention threshold $\omega$.
\ENSURE Robust model parameters $\theta^*$
\STATE Initialize network parameters $\theta$
\FOR{each epoch}
    \FOR{each mini-batch $(x, y)$}
        \STATE // Generate adversarial examples
        \STATE $\hat{x} \leftarrow \arg\max_{||x'-x||_p \leq \epsilon} \mathcal{L}_{AT}(f_\theta(x'), y)$
        \STATE $\hat{z} \leftarrow f_\theta(\hat{x})$
        
        \STATE // Generate inverse adversarial examples
        \STATE $\check{x} \leftarrow \arg\min_{||x'-x||_p \leq \epsilon} \mathcal{L}_{Inv}(f_\theta(x'), y)$
        \STATE $\check{z} \leftarrow f_\theta(\check{x})$
        
        \STATE // Compute attention maps using a selected method $\mathcal{A} \in \{\text{Grad-CAM, Integrared-Grad, SAM, \textit{etc.}}\}$
        \STATE $M \leftarrow \mathcal{A}(x)$ \COMMENT{Different attention map computation techniques}
        
        \STATE // Extract background features from inverse adversarial examples
        \STATE $[\check{x}_{(B)}]_{(i,j)} \leftarrow \mathbb{I}_{(M_{i,j} < \omega)} \cdot \check{x}_{(i,j)}$
        \STATE $\check{z}_{(B)} \leftarrow f_\theta(\check{x}_{(B)})$
        
        \STATE // Compute debiased high-confidence logits
        \STATE $\check{z}^* \leftarrow \check{z} - \check{z}_{(B)}$
        
        \STATE // Compute DHLR loss
        \STATE $\mathcal{L}_{DHLR} \leftarrow \mathcal{L}_{KL}(\phi(\check{z}^*) || \phi(\hat{z}))$
        
        \STATE // Compute FLOE loss
        \STATE $\mathcal{L}_{FLOE} \leftarrow -|\check{z} - \frac{\check{z} \cdot \check{z}_{(B)}}{|\check{z}_{(B)}|^2} \cdot \check{z}_{(B)}|_p$
        
        \STATE // Compute total loss
        \STATE $\mathcal{L}_{DHAT} \leftarrow \mathcal{L}_{AT}(\hat{z}, y) + \lambda_1 \cdot \mathcal{L}_{DHLR} + \lambda_2 \cdot \mathcal{L}_{FLOE}$
        
        \STATE // Update network parameters
        \STATE $\theta^* \leftarrow \theta - \nabla_\theta \mathcal{L}_{DHAT}$
    \ENDFOR
\ENDFOR
\RETURN $\theta^*$
\end{algorithmic}
\end{algorithm}
Algorithm~\ref{alg:dhat} outlines the complete workflow of our proposed \textit{Debiased High-Confidence Adversarial Training} (DHAT) framework, which systematically mitigates spurious correlations in adversarial training through two primary components: Debiased High-Confidence Logit Regularization (DHLR) and Foreground Logit Orthogonal Enhancement (FLOE).

To efficiently compute attention maps, we employ Grad-CAM (Line 11) due to its computational efficiency. However, our framework is compatible with alternative saliency-based methods, such as SAM~\cite{kirillov2023segment}, which can provide enhanced performance at a higher computational cost. The extraction of background features (Lines 13-14) is performed using an adaptive threshold $\omega$, ensuring the identification of non-discriminative regions that contribute to spurious correlations. The overall training objective (Line 23) integrates standard adversarial training with our proposed debiasing regularization terms, weighted by  $\lambda_1$ and $\lambda_2$.

Despite incorporating additional minimal computational steps, the DHAT framework introduces only a marginal increase in resource consumption compared to standard adversarial training while significantly enhancing model robustness. Crucially, DHAT achieves these improvements without compromising clean-data accuracy, making it a highly practical and scalable solution for real-world adversarial defense applications.

\input{Table/transfer_attack}

\section{Performance Under Transfer Attack}
\label{sec:trans_task}
In this part, we evaluate our proposed model's performance under transfer attacks and compare it with the UIAT~\cite{UIAT}. 
Transfer attacks evaluate a model's robustness by testing its performance against adversarial examples generated from different source models, which simulate real-world conditions where attackers may use varied strategies. This testing is essential to ensure that a model's adversarial defenses generalize effectively beyond its training environment and are robust against diverse attack methods. 
By evaluating the performance under transfer attacks, we ensure that our model is robust not only to attacks generated by its own architecture but also to those from different models, providing a more comprehensive measure of robustness.

Table~\ref{tb:trans_attack} presents the robust accuracy of various models under transfer attacks with an $\ell_\infty$-norm perturbation of $\epsilon = 8$. The table compares the performance of the UIAT method with that of our proposed model across different target models and transfer attack types.

\subsection{Robustness Performance}
Our model consistently outperforms UIAT across all transfer adversarial attacks and target models. 
For example, when the WRN28-10 model is trained with our proposed DHAT and subjected to AA attacks generated from ResNet-50, VGG16, and Inc-V3 source models, the defense success rate improves by 1.21\%, 1.14\%, and 1.44\% compared to UIAT, respectively. 
This demonstrates the model's robustness, highlighting its effectiveness not only in specific attack scenarios but also across various model architectures.

\subsection{Robustness Across Various Attack Types}
Our proposed model DHAT demonstrates enhanced robustness, particularly against more challenging attacks such as PGD-50 and C\&W. For instance, when using the ResNet-18 as the source model, our method achieves improvements of up to 1.88\% and 2.39\% under PGD-50 and C\&W attacks from WRN28-10, respectively, compared to UIAT. This highlights the model's superior capability to withstand adversarial perturbations.

\subsection{Generalization Across Various Source Models}
Our model exhibits strong performance across a range of source models, including VGG-16, WRN28-10, and ResNet-18. This indicates that the improved robustness of our model is not limited to specific attacks generated from source architectures but generalizes effectively across different adversarial settings.

\input{Table/Additional_data}

\section{Performance with Generated Data} \label{sec:generated}
We employ Diffusion Denoising Probabilistic Models (DDPM)~\cite{ho2020denoising} to generate an additional 50K samples for both CIFAR-10 and CIFAR-100 datasets following the \cite{wang2023better}. This synthetic data is then used to augment the training of both our method and all baseline models. The performance comparisons, as shown in \Cref{st:DDPM_data}, illustrate the impact of additional data on robustness and generalization.

\subsection{Robustness Performance}
The results in \Cref{st:DDPM_data} demonstrate that our method outperforms baseline methods in terms of robustness when trained with the additional synthetic data. This improvement highlights the effectiveness of our approach in leveraging extra data to enhance model robustness, particularly under adversarial conditions.

\subsection{Generalization Performance}
We observe that, compared to the results in Table 1, most models trained with the additional data exhibit a reduced Robust Gap, indicating improved generalization. However, both UIAT and SGLR exhibit limited improvements in robust generalization. These methods rely on spurious correlations during training, which hinders their generalization performance even with the added data. Although UIAT and SGLR benefit from enhanced robustness due to increased data diversity, their robust generalization remains suboptimal, likely due to their dependence on non-essential features. This finding underscores the unique advantage of our method in achieving both robust accuracy and generalization with enriched datasets.

\begin{table}[t]
    \begin{center}
    \caption{Comparison of robustness~(\%) and robust generalization gap~(\%) for using various attention map generation techniques using WRN28-10 on the CIFAR-10.}
    \label{st:sam}
    \resizebox{\linewidth}{!}
    {
        \begin{tabular}{lccccc}
        \toprule[1.5pt] 
        Method & Clean$\uparrow$ & PGD-10$\uparrow$ & C\&W$\uparrow$ &AA$\uparrow$&Robust Gap$\downarrow$ \\
        \midrule[0.4pt]
        -&82.94&58.66&54.11&52.17&7.92\\
        Grad-CAM~\cite{grad_cam_paper} &83.95&60.49&55.27&53.10&3.51\\
        Integrated-Grad~\cite{sundararajan2017axiomatic} &83.97&60.35&55.18&53.04&3.68\\
        SOLO~\cite{wang2020solo} &84.26&61.60&56.74&54.93&3.09\\
        SAM~\cite{kirillov2023segment} &85.65&62.44&58.10&56.38&2.46\\
        \bottomrule[1.5pt]
        \end{tabular}
   }
    \end{center}
\end{table}

\section{Impact of Foreground-Background Recognition Techniques}
\label{sec:recognition}
In this part, we investigate the influence of different attention map generation techniques on the performance of Debiased High-Confidence Logit Regularization (DHAT), as seen in \Cref{st:sam}. Our exploration extends beyond the initially employed Grad-CAM method to encompass a variety of attention map generation approaches, thereby providing a comprehensive analysis of their effects on model robustness and generalization.

\begin{figure*}[t]
    \centering
    \begin{tabular}{@{}c@{\hspace{-0.25mm}}c@{}@{\hspace{-0.25mm}}c@{}@{\hspace{-0.25mm}}c@{}}
        \includegraphics[width=0.245\linewidth]{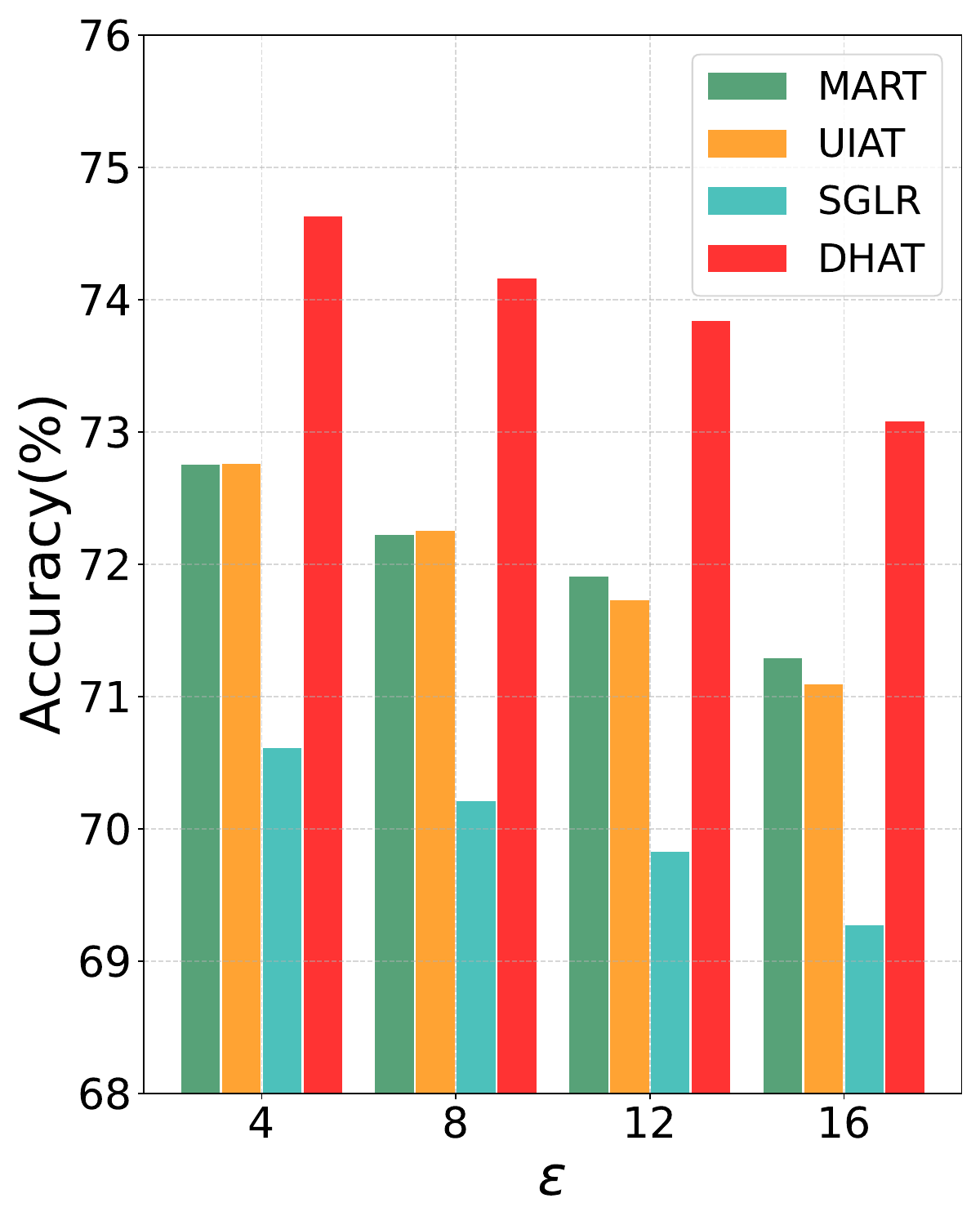}&
        \includegraphics[width=0.245\linewidth]{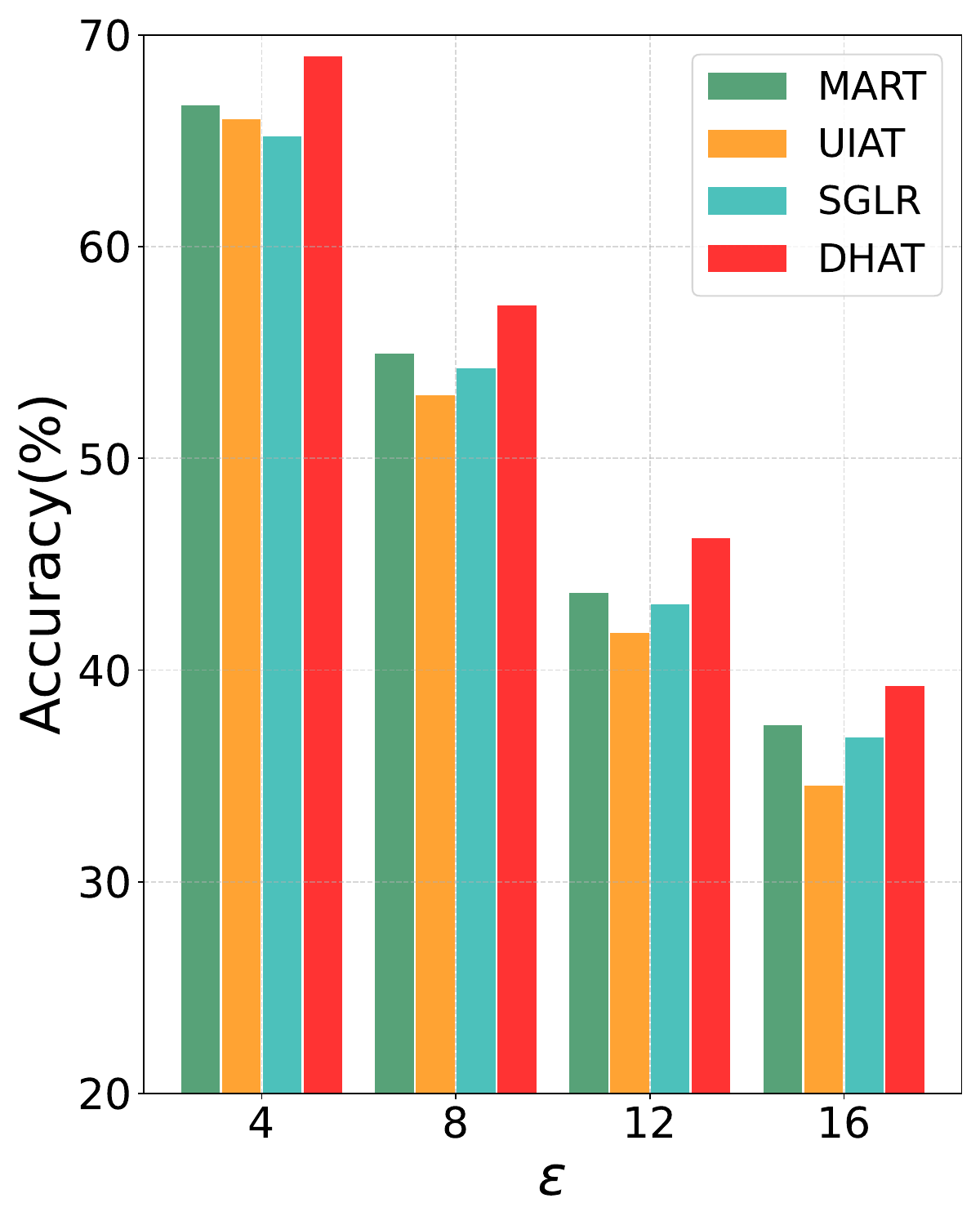}
        &
        \includegraphics[width=0.245\linewidth]{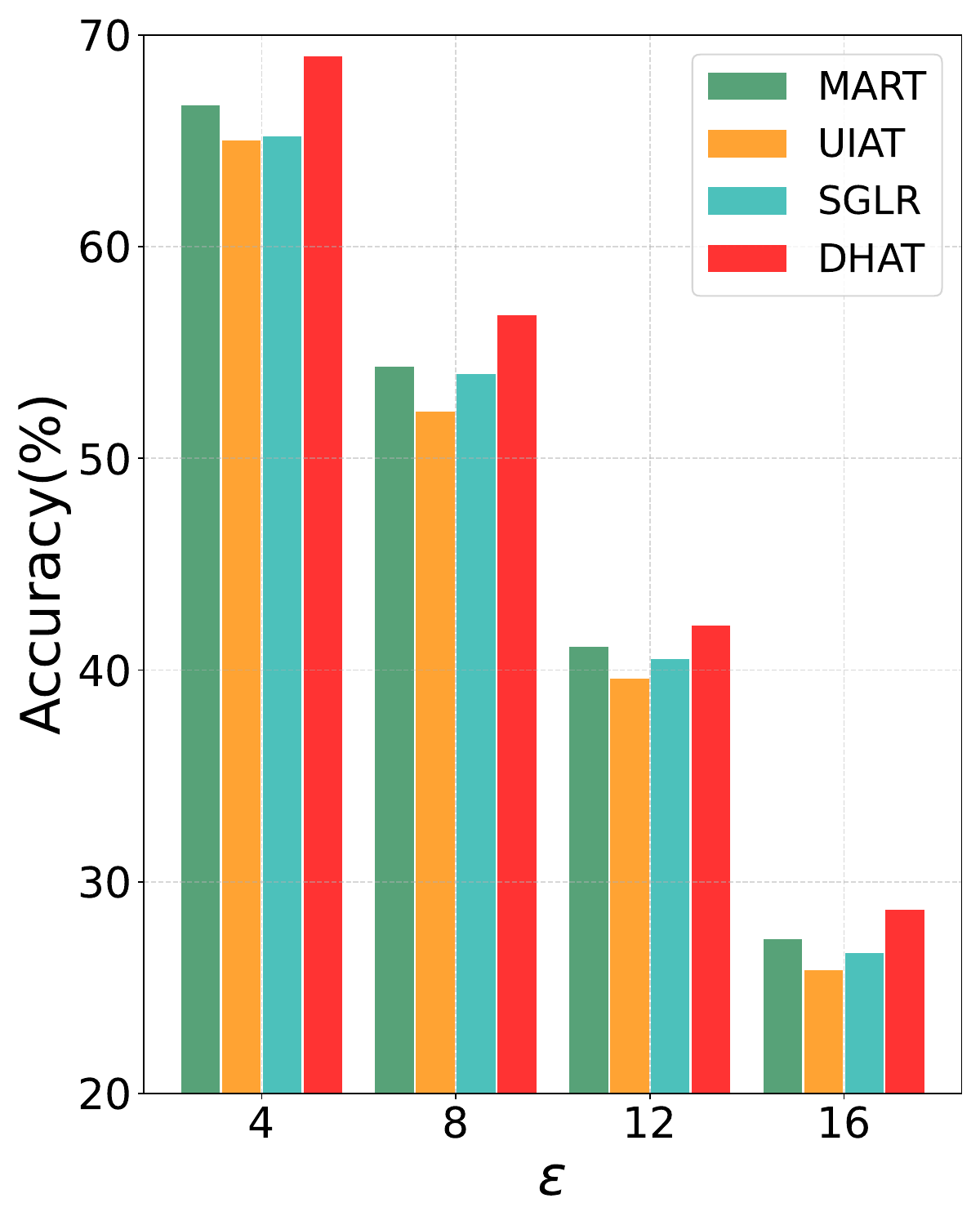}
        &
        \includegraphics[width=0.245\linewidth]{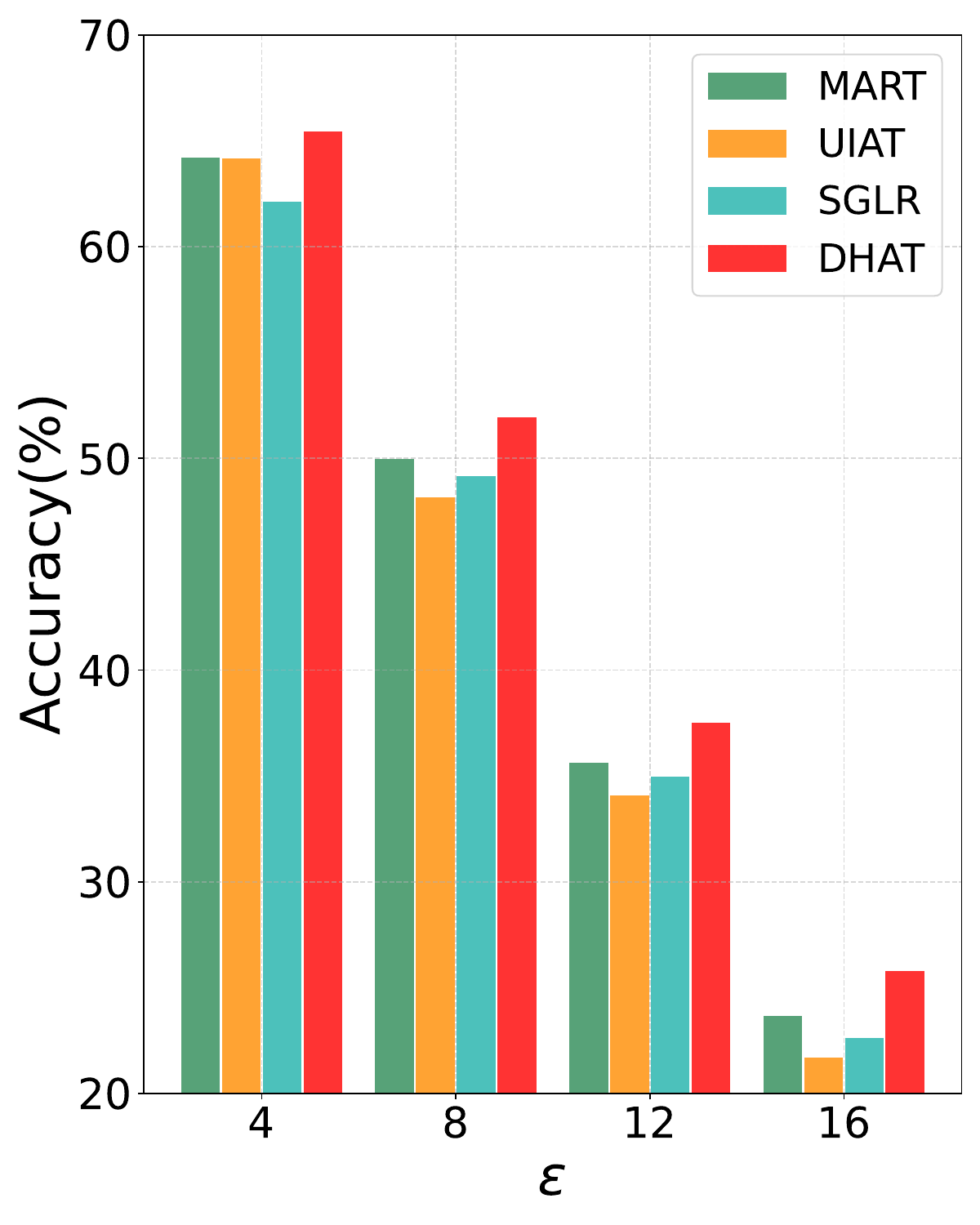}\\
        \parbox{0.245\linewidth}{\centering \footnotesize{\ \ \ \ (a) FGSM Attack}} & 
        \parbox{0.245\linewidth}{\centering \footnotesize{\ \ \ \ (b) PGD-10 Attack}}& 
        \parbox{0.245\linewidth}{\centering \footnotesize{\ \ \ \ (c) PGD-20 Attack}}& 
        \parbox{0.245\linewidth}{\centering \footnotesize{\ \ \ \ (d) C\&W Attack}}
    \end{tabular}
    \vspace{-0.5em}
    \caption{Comparisons with varying $\epsilon$ values using ResNet-18 on the CIFAR-10. The $x$-axis represents the $\epsilon$ value, while $y$-axis represents the robust accuracy~(\%).}
    \label{fig:Epsilon_4}
\end{figure*}

\subsection{Exploration of Various Attention Map}
The primary approach detailed in the main text utilizes Grad-CAM~\cite{grad_cam_paper}, a widely adopted method in weakly supervised object segmentation~\cite{liu2022practical}, to extract foreground and background feature maps efficiently. Grad-CAM offers a balance between computational efficiency and effectiveness, making it suitable for large-scale evaluations. However, to understand the broader applicability and potential improvements, we incorporated additional attention map generation techniques, including Integrated Gradients~\cite{sundararajan2017axiomatic}, SOLO~\cite{wang2020solo}, and SAM~\cite{kirillov2023segment}.

\subsection{Evaluation of Advanced Attention Map Models}
Incorporating more sophisticated attention map generation models, such as SAM, demonstrated enhanced robustness and reduced robust generalization gaps compared to simpler methods like Grad-CAM. These advanced models provide finer and more accurate segmentation of foreground and background features, which in turn leads to better alignment of high-confidence logits under adversarial conditions. However, this improvement in performance comes at the cost of increased computational overhead. More complex models require greater processing power and longer training times, which may be a limiting factor in resource-constrained environments.

\subsection{Trade-Off Between Performance and Computational Efficiency}
While the adoption of advanced attention map techniques can yield superior performance metrics, researchers must consider the trade-offs involved. Enhanced models may offer marginal gains in robustness and generalization, but the additional computational resources and time required may not always justify these benefits, especially in applications where real-time processing is essential. Therefore, the choice of attention map generation technique should be informed by the specific requirements and constraints of the deployment scenario.

\section{Robustness under Various Magnitude Attacks}
\label{sec:epsilon}
In this part, we evaluate the robustness of our proposed model under various levels of adversarial perturbations, quantified by different $\epsilon$ values. 
We compare our model with two baseline methods, MART~\cite{MART} and UIAT~\cite{UIAT}, and the state-of-the-art SGLR method~\cite{SGLR} across various attack types, including FGSM, PGD-10, PGD-20, and C\&W. 
We aim to demonstrate the superior generalization and robustness of our model under progressively challenging adversarial conditions. 
\Cref{fig:Epsilon_4} shows the accuracy of different methods under various $\epsilon$ values for each attack type.

The comparative analysis of our model with MART, UIAT, and SGLR reveals several key insights:
\subsection{Robustness Across Various Attack Types}
Our model consistently outperforms the baselines and SGLR across all attack types and $\epsilon$ values. This consistent superiority underscores our model's robust generalization ability, allowing it to maintain high accuracy even under more challenging adversarial attack conditions.
\subsection{Generalization Against Various Attacks} 
The results suggest that the defensive mechanisms in our model are highly effective in mitigating the impact of various adversarial attacks (\emph{i.e.,} FGSM, PGD-10, PGD-20, C\&W). This robustness is especially evident in scenarios with stronger attacks and higher $\epsilon$ values, where our model shows superior performance compared to existing methods.
\subsection{Practical Applicability} 
The enhanced robustness of our model across different $\epsilon$ values and attack types demonstrates its practical applicability in real-world scenarios, where adversarial perturbations can vary in strength and sophistication.

%% file: Table/transfer_attack.tex
\begin{table*}[t]
    \begin{center}
    \caption{Transfer attack accuracy~(\%) in the single-model transfer scenario. The \textbf{number} in bold indicates the best accuracy.}
    \resizebox{\textwidth}{!}
    {
    \begin{tabular}{c|ccc|ccc|ccc}
        \toprule
        \multirow{3}*{Attack ($\epsilon=8$)} & \multicolumn{9}{c}{Performance of UIAT~/~DHAT}\\
        \cmidrule{2-10}
         & \multicolumn{3}{c|}{Target: VGG-16} & \multicolumn{3}{c|}{Target: WRN28-10} & \multicolumn{3}{c}{Target: ResNet-18}\\
        & $\Rightarrow$ ResNet-50 &$\Rightarrow$ Inc-V3 & $\Rightarrow$ WRN28-10&$ \Rightarrow$ ResNet-50 &$\Rightarrow$ VGG16 & $\Rightarrow$ Inc-v3&$ \Rightarrow$ WRN28-10 &$\Rightarrow$ VGG16 & $\Rightarrow$ Inc-v3\\ \midrule
        FGSM    &63.42/\textbf{64.83}&63.44/\textbf{64.51}&63.73/\textbf{64.88}&77.11/\textbf{78.73}&78.08/\textbf{79.27}&77.84/\textbf{78.69}&73.01/\textbf{74.80}&74.00/\textbf{75.03}&72.43/\textbf{73.18}\\
        PGD-10  &53.09/\textbf{54.46}&52.81/\textbf{53.96}&54.17/\textbf{55.36}&61.93/\textbf{62.87}&64.73/\textbf{65.70}&61.40/\textbf{62.56}&58.56/\textbf{60.30}&59.91/\textbf{60.89}&57.82/\textbf{58.75}\\
        PGD-20  &52.73/\textbf{54.25}&52.63/\textbf{53.70}&53.80/\textbf{55.29}&61.52/\textbf{62.68}&64.60/\textbf{65.65}&61.19/\textbf{62.38}&58.13/\textbf{60.04}&59.80/\textbf{60.78}&57.48/\textbf{58.42}\\
        PGD-50  &52.68/\textbf{54.25}&52.62/\textbf{53.67}&53.77/\textbf{55.27}&61.51/\textbf{62.65}&64.56/\textbf{65.68}&61.17/\textbf{62.24}&58.15/\textbf{60.03}&59.78/\textbf{60.83}&57.45/\textbf{58.46}\\
        C\&W    &51.90/\textbf{53.44}&51.38/\textbf{52.61}&55.31/\textbf{56.35}&61.05/\textbf{62.02}&63.32/\textbf{64.40}&60.76/\textbf{61.98}&57.73/\textbf{60.12}&57.44/\textbf{58.74}&57.71/\textbf{59.25}\\
        AA      &56.31/\textbf{57.94}&56.02/\textbf{56.98}&59.65/\textbf{60.77}&67.18/\textbf{68.39}&71.86/\textbf{73.00}&64.39/\textbf{65.83}&62.36/\textbf{63.19}&65.19/\textbf{66.07}&60.69/\textbf{61.94}\\
        \bottomrule
    \end{tabular}
    }
    \label{tb:trans_attack}
    \end{center}
\end{table*}

%% file: Table/Additional_data.tex
\begin{table}[t]
        \begin{center}
        \caption{Comparison of robustness~(\%) and robust generalization gap~(\%) for models trained on generated data. The \textbf{bolded numbers} indicate the best performance.}
        \resizebox{\linewidth}{!}
        {
            \begin{tabular}{lcccccc}
            \toprule[1.5pt] 
            \textbf{CIFAR-10} & \multicolumn{3}{c}{ResNet-18} &\multicolumn{3}{c}{WRN28-10}\\
            \cmidrule(lr){2-4}  \cmidrule(lr){5-7} 
            Method&Clean$\uparrow$&AA$\uparrow$&Robust Gap$\downarrow$&Clean$\uparrow$&AA$\uparrow$&Robust Gap$\downarrow$\\ 
            \midrule
            MART~\cite{MART}     &83.45&49.45&2.91&84.26&51.95&8.73\\
            AWP~\cite{AWP}       &83.78&50.79&2.03&84.10&53.29&6.11\\
            FSR~\cite{FSR}       &83.19&50.53&2.35&83.88&53.03&7.05\\
            CFA~\cite{CFA}       &84.97&50.85&2.98&85.81&53.35&8.94\\
            UIAT~\cite{UIAT}     &85.10&52.09&3.04&86.73&54.59&9.41\\
            SGLR~\cite{SGLR}     &86.35&51.27&2.75&87.72&53.77&11.25\\
            \rowcolor{gray!20}
            DHAT~(Ours)          &\textbf{87.62}&\textbf{54.42}&\textbf{0.83}&\textbf{88.94}&\textbf{56.92}&\textbf{2.48}\\
            \midrule[1.5pt]
            \textbf{CIFAR-100} & \multicolumn{3}{c}{ResNet-18} &\multicolumn{3}{c}{WRN28-10}\\
            \cmidrule(lr){2-4}  \cmidrule(lr){5-7} 
            Method&Clean$\uparrow$&AA$\uparrow$&Robust Gap$\downarrow$&Clean$\uparrow$&AA$\uparrow$&Robust Gap$\downarrow$\\ 
            \midrule
            MART~\cite{MART}     &54.73&27.70&2.88&55.87&30.20&8.86\\
            AWP~\cite{AWP}       &57.55&29.33&2.37&59.71&31.83&7.13\\
            FSR~\cite{FSR}       &58.10&28.94&2.47&59.03&30.44&7.40\\
            CFA~\cite{CFA}       &60.13&28.85&3.01&61.56&29.61&9.13\\
            UIAT~\cite{UIAT}     &59.92&28.48&4.47&60.24&30.98&14.32\\
            SGLR~\cite{SGLR}     &61.25&29.10&4.15&62.39&30.15&17.05\\
            \rowcolor{gray!20}
            DHAT~(Ours)          &\textbf{63.14}&\textbf{32.21}&\textbf{0.98}&\textbf{64.80}&\textbf{34.71}&\textbf{2.96}\\
            \bottomrule[1.5pt]
            \end{tabular}
        }
        \label{st:DDPM_data}
        \end{center}
\end{table}

%% file: main.bbl
\begin{thebibliography}{55}
\providecommand{\natexlab}[1]{#1}
\providecommand{\url}[1]{\texttt{#1}}
\expandafter\ifx\csname urlstyle\endcsname\relax
  \providecommand{\doi}[1]{doi: #1}\else
  \providecommand{\doi}{doi: \begingroup \urlstyle{rm}\Url}\fi

\bibitem[Ahmad et~al.(2024)Ahmad, B{\'e}reux, Baret, Hashemi, and
  Lecue]{ahmad2024causal}
Ola Ahmad, Nicolas B{\'e}reux, Lo{\"\i}c Baret, Vahid Hashemi, and Freddy
  Lecue.
\newblock Causal analysis for robust interpretability of neural networks.
\newblock In \emph{WAVC}, pages 4685--4694, 2024.

\bibitem[Andriushchenko et~al.(2020)Andriushchenko, Croce, Flammarion, and
  Hein]{Square}
Maksym Andriushchenko, Francesco Croce, Nicolas Flammarion, and Matthias Hein.
\newblock Square attack: a query-efficient black-box adversarial attack via
  random search.
\newblock In \emph{ECCV}, 2020.

\bibitem[Asgari et~al.(2022)Asgari, Khani, Khani, Gholami, Tran, Mahdavi~Amiri,
  and Hamarneh]{asgari2022masktune}
Saeid Asgari, Aliasghar Khani, Fereshte Khani, Ali Gholami, Linh Tran, Ali
  Mahdavi~Amiri, and Ghassan Hamarneh.
\newblock Masktune: Mitigating spurious correlations by forcing to explore.
\newblock In \emph{NeurIPS}, 2022.

\bibitem[Carlini and Wagner(2017)]{CW}
Nicholas Carlini and David Wagner.
\newblock Towards evaluating the robustness of neural networks.
\newblock \emph{IEEE Symp. Security Privacy}, pages 39--57, 2017.

\bibitem[Cho et~al.(2023)Cho, Han, and Kim]{cho2023anti}
Hyuna Cho, Yubin Han, and Won~Hwa Kim.
\newblock Anti-adversarial consistency regularization for data augmentation:
  Applications to robust medical image segmentation.
\newblock In \emph{MICAI}, 2023.

\bibitem[Crabb{\'e} and van~der Schaar(2023)]{crabbe2023evaluating}
Jonathan Crabb{\'e} and Mihaela van~der Schaar.
\newblock Evaluating the robustness of interpretability methods through
  explanation invariance and equivariance.
\newblock In \emph{NeurIPS}, 2023.

\bibitem[Croce and Hein(2020{\natexlab{a}})]{AA}
Francesco Croce and Matthias Hein.
\newblock Reliable evaluation of adversarial robustness with an ensemble of
  diverse parameter-free attacks.
\newblock In \emph{ICML}, 2020{\natexlab{a}}.

\bibitem[Croce and Hein(2020{\natexlab{b}})]{FAB}
Francesco Croce and Matthias Hein.
\newblock Minimally distorted adversarial examples with a fast adaptive
  boundary attack.
\newblock In \emph{ICML}, 2020{\natexlab{b}}.

\bibitem[Dong et~al.(2023)Dong, Moosavi-Dezfooli, Lai, and Xie]{UIAT}
Junhao Dong, Seyed-Mohsen Moosavi-Dezfooli, Jianhuang Lai, and Xiaohua Xie.
\newblock The enemy of my enemy is my friend: Exploring inverse adversaries for
  improving aversarial training.
\newblock In \emph{CVPR}, 2023.

\bibitem[Dong et~al.(2020)Dong, Han, Chen, Liu, Bian, Ma, Li, Wang, Zhang, and
  Yu]{dong2020robust}
Xiaoyi Dong, Jiangfan Han, Dongdong Chen, Jiayang Liu, Huanyu Bian, Zehua Ma,
  Hongsheng Li, Xiaogang Wang, Weiming Zhang, and Nenghai Yu.
\newblock Robust superpixel-guided attentional adversarial attack.
\newblock In \emph{CVPR}, 2020.

\bibitem[Fawzi et~al.(2018)Fawzi, Fawzi, and Fawzi]{fawzi2018adversarial}
Alhussein Fawzi, Hamza Fawzi, and Omar Fawzi.
\newblock Adversarial vulnerability for any classifier.
\newblock In \emph{NeurIPS}, 2018.

\bibitem[Fowl et~al.(2021)Fowl, Goldblum, Chiang, Geiping, Czaja, and
  Goldstein]{fowl2021adversarial}
Liam Fowl, Micah Goldblum, Ping-yeh Chiang, Jonas Geiping, Wojciech Czaja, and
  Tom Goldstein.
\newblock Adversarial examples make strong poisons.
\newblock In \emph{NeurIPS}, 2021.

\bibitem[Gao et~al.(2024)Gao, Chen, Chen, Wang, and Lu]{gao2024avsegformer}
Shengyi Gao, Zhe Chen, Guo Chen, Wenhai Wang, and Tong Lu.
\newblock Avsegformer: Audio-visual segmentation with transformer.
\newblock In \emph{AAAI}, 2024.

\bibitem[Goodfellow et~al.(2014)Goodfellow, Shlens, and Szegedy]{FGSM}
Ian~J Goodfellow, Jonathon Shlens, and Christian Szegedy.
\newblock Explaining and harnessing adversarial examples.
\newblock \emph{arXiv preprint arXiv:1412.6572}, 2014.

\bibitem[Han et~al.(2023)Han, Wang, Su, Huang, and Tian]{Bias_corre_2023}
Xinzhe Han, Shuhui Wang, Chi Su, Qingming Huang, and Qi Tian.
\newblock General greedy de-bias learning.
\newblock \emph{IEEE TPAMI}, 45\penalty0 (8):\penalty0 9789--9805, 2023.

\bibitem[He et~al.(2016)He, Zhang, Ren, and Sun]{ResNet}
Kaiming He, Xiangyu Zhang, Shaoqing Ren, and Jian Sun.
\newblock Deep residual learning for image recognition.
\newblock In \emph{CVPR}, 2016.

\bibitem[Ho et~al.(2020)Ho, Jain, and Abbeel]{ho2020denoising}
Jonathan Ho, Ajay Jain, and Pieter Abbeel.
\newblock Denoising diffusion probabilistic models.
\newblock \emph{NeurIPS}, 2020.

\bibitem[Howard(2019)]{Howard_Imagenette_2019}
FastAI~Jeremy Howard.
\newblock Imagenette: A smaller subset of imagenet.
\newblock In \emph{Github}. https://github.com/fastai/imagenette, 2019.

\bibitem[Huang et~al.(2023)Huang, Fan, Liu, Zhang, Zhang, Salzmann,
  S{\"u}sstrunk, and Wang]{huang2023fast}
Zhichao Huang, Yanbo Fan, Chen Liu, Weizhong Zhang, Yong Zhang, Mathieu
  Salzmann, Sabine S{\"u}sstrunk, and Jue Wang.
\newblock Fast adversarial training with adaptive step size.
\newblock \emph{IEEE TIP}, 32:\penalty0 6102--6114, 2023.

\bibitem[Izmailov et~al.(2022)Izmailov, Kirichenko, Gruver, and
  Wilson]{izmailov2022feature}
Pavel Izmailov, Polina Kirichenko, Nate Gruver, and Andrew~G Wilson.
\newblock On feature learning in the presence of spurious correlations.
\newblock In \emph{NeurIPS}, 2022.

\bibitem[Kim et~al.(2023)Kim, Cho, Jung, and Yoon]{FSR}
Woo~Jae Kim, Yoonki Cho, Junsik Jung, and Sung-Eui Yoon.
\newblock Feature separation and recalibration for adversarial robustness.
\newblock In \emph{CVPR}, 2023.

\bibitem[Kirillov et~al.(2023)Kirillov, Mintun, Ravi, Mao, Rolland, Gustafson,
  Xiao, Whitehead, Berg, Lo, et~al.]{kirillov2023segment}
Alexander Kirillov, Eric Mintun, Nikhila Ravi, Hanzi Mao, Chloe Rolland, Laura
  Gustafson, Tete Xiao, Spencer Whitehead, Alexander~C Berg, Wan-Yen Lo, et~al.
\newblock Segment anything.
\newblock In \emph{ICCV}, 2023.

\bibitem[Krizhevsky et~al.(2009)Krizhevsky, Hinton, et~al.]{CIFAR}
Alex Krizhevsky, Geoffrey Hinton, et~al.
\newblock Learning multiple layers of features from tiny images.
\newblock In \emph{Toronto, ON, Canada}, 2009.

\bibitem[Li and Liu(2023)]{li2023wat}
Boqi Li and Weiwei Liu.
\newblock Wat: improve the worst-class robustness in adversarial training.
\newblock In \emph{AAAI}, 2023.

\bibitem[Li et~al.(2024{\natexlab{a}})Li, Xiao, and Tang]{li2024asam}
Bo Li, Haoke Xiao, and Lv Tang.
\newblock Asam: Boosting segment anything model with adversarial tuning.
\newblock In \emph{CVPR}, 2024{\natexlab{a}}.

\bibitem[Li et~al.(2024{\natexlab{b}})Li, Jiang, Zhang, Ren, Liu, Zou, Xu, Li,
  Yang, Li, et~al.]{li2024visual}
Feng Li, Qing Jiang, Hao Zhang, Tianhe Ren, Shilong Liu, Xueyan Zou, Huaizhe
  Xu, Hongyang Li, Jianwei Yang, Chunyuan Li, et~al.
\newblock Visual in-context prompting.
\newblock In \emph{CVPR}, 2024{\natexlab{b}}.

\bibitem[Li et~al.(2024{\natexlab{c}})Li, Yu, Wei, Jin, Zhang, and Chan]{SGLR}
Zhuorong Li, Daiwei Yu, Lina Wei, Canghong Jin, Yun Zhang, and Sixian Chan.
\newblock Soften to defend: Towards adversarial robustness via self-guided
  label refinement.
\newblock In \emph{CVPR}, 2024{\natexlab{c}}.

\bibitem[Liu et~al.(2022)Liu, Cheng, Gao, Liu, Zhang, and
  Song]{liu2022practical}
Ye Liu, Yaya Cheng, Lianli Gao, Xianglong Liu, Qilong Zhang, and Jingkuan Song.
\newblock Practical evaluation of adversarial robustness via adaptive auto
  attack.
\newblock In \emph{CVPR}, 2022.

\bibitem[Madry et~al.(2018)Madry, Makelov, Schmidt, Tsipras, and
  Vladu]{PGD_Attack}
Aleksander Madry, Aleksandar Makelov, Ludwig Schmidt, Dimitris Tsipras, and
  Adrian Vladu.
\newblock Towards deep learning models resistant to adversarial attacks.
\newblock In \emph{ICLR}, 2018.

\bibitem[Ming et~al.(2022)Ming, Yin, and Li]{ming2022impact}
Yifei Ming, Hang Yin, and Yixuan Li.
\newblock On the impact of spurious correlation for out-of-distribution
  detection.
\newblock In \emph{AAAI}, 2022.

\bibitem[Russakovsky et~al.(2015)Russakovsky, Deng, Su, Krause, Satheesh, Ma,
  Huang, Karpathy, Khosla, Bernstein, et~al.]{russakovsky2015imagenet}
Olga Russakovsky, Jia Deng, Hao Su, Jonathan Krause, Sanjeev Satheesh, Sean Ma,
  Zhiheng Huang, Andrej Karpathy, Aditya Khosla, Michael Bernstein, et~al.
\newblock Imagenet large scale visual recognition challenge.
\newblock \emph{IJCV}, 115:\penalty0 211--252, 2015.

\bibitem[Salman et~al.(2021)Salman, Ilyas, Engstrom, Vemprala, Madry, and
  Kapoor]{salman2021unadversarial}
Hadi Salman, Andrew Ilyas, Logan Engstrom, Sai Vemprala, Aleksander Madry, and
  Ashish Kapoor.
\newblock Unadversarial examples: Designing objects for robust vision.
\newblock In \emph{NeurIPS}, 2021.

\bibitem[Selvaraju et~al.(2017{\natexlab{a}})Selvaraju, Cogswell, Das,
  Vedantam, Parikh, and Batra]{grad_cam_paper}
Ramprasaath~R Selvaraju, Michael Cogswell, Abhishek Das, Ramakrishna Vedantam,
  Devi Parikh, and Dhruv Batra.
\newblock Grad-cam: Visual explanations from deep networks via gradient-based
  localization.
\newblock In \emph{ICCV}, 2017{\natexlab{a}}.

\bibitem[Selvaraju et~al.(2017{\natexlab{b}})Selvaraju, Cogswell, Das,
  Vedantam, Parikh, and Batra]{selvaraju2017grad}
Ramprasaath~R Selvaraju, Michael Cogswell, Abhishek Das, Ramakrishna Vedantam,
  Devi Parikh, and Dhruv Batra.
\newblock Grad-cam: Visual explanations from deep networks via gradient-based
  localization.
\newblock In \emph{CVPR}, 2017{\natexlab{b}}.

\bibitem[Seo et~al.(2022)Seo, Lee, and Han]{seo2022information}
Seonguk Seo, Joon-Young Lee, and Bohyung Han.
\newblock Information-theoretic bias reduction via causal view of spurious
  correlation.
\newblock In \emph{AAAI}, 2022.

\bibitem[Shafahi et~al.(2019)Shafahi, Najibi, Ghiasi, Xu, Dickerson, Studer,
  Davis, Taylor, and Goldstein]{shafahi2019adversarial}
Ali Shafahi, Mahyar Najibi, Mohammad~Amin Ghiasi, Zheng Xu, John Dickerson,
  Christoph Studer, Larry~S Davis, Gavin Taylor, and Tom Goldstein.
\newblock Adversarial training for free!
\newblock In \emph{NeurIPS}, 2019.

\bibitem[Shao et~al.(2025)Shao, Tao, Qin, You, Sui, and Wang]{shao2025holitom}
Kele Shao, Keda Tao, Can Qin, Haoxuan You, Yang Sui, and Huan Wang.
\newblock Holitom: Holistic token merging for fast video large language models.
\newblock \emph{arXiv preprint arXiv:2505.21334}, 2025.

\bibitem[Simonyan and Zisserman(2014)]{VGG}
Karen Simonyan and Andrew Zisserman.
\newblock Very deep convolutional networks for large-scale image recognition.
\newblock \emph{arXiv preprint arXiv:1409.1556}, 2014.

\bibitem[Singh et~al.(2023)Singh, Croce, and Hein]{singh2023revisiting}
Naman~Deep Singh, Francesco Croce, and Matthias Hein.
\newblock Revisiting adversarial training for imagenet: Architectures, training
  and generalization across threat models.
\newblock In \emph{NeurIPS}, 2023.

\bibitem[Subramanian et~al.(2024)Subramanian, Sizikova, Majaj, and
  Pelli]{subramanian2024spatial}
Ajay Subramanian, Elena Sizikova, Najib Majaj, and Denis Pelli.
\newblock Spatial-frequency channels, shape bias, and adversarial robustness.
\newblock In \emph{NeurIPS}, 2024.

\bibitem[Sundararajan et~al.(2017)Sundararajan, Taly, and
  Yan]{sundararajan2017axiomatic}
Mukund Sundararajan, Ankur Taly, and Qiqi Yan.
\newblock Axiomatic attribution for deep networks.
\newblock In \emph{ICML}, 2017.

\bibitem[Szegedy et~al.(2014)Szegedy, Zaremba, Sutskever, Bruna, Erhan,
  Goodfellow, and Fergus]{szegedy2013intriguing}
Christian Szegedy, Wojciech Zaremba, Ilya Sutskever, Joan Bruna, Dumitru Erhan,
  Ian Goodfellow, and Rob Fergus.
\newblock Intriguing properties of neural networks.
\newblock In \emph{ICLR}, 2014.

\bibitem[Szegedy et~al.(2015)Szegedy, Liu, Jia, Sermanet, Reed, Anguelov,
  Erhan, Vanhoucke, and Rabinovich]{Inception_NET}
Christian Szegedy, Wei Liu, Yangqing Jia, Pierre Sermanet, Scott Reed, Dragomir
  Anguelov, Dumitru Erhan, Vincent Vanhoucke, and Andrew Rabinovich.
\newblock Going deeper with convolutions.
\newblock In \emph{CVPR}, 2015.

\bibitem[Wang et~al.(2020{\natexlab{a}})Wang, Kong, Shen, Jiang, and
  Li]{wang2020solo}
Xinlong Wang, Tao Kong, Chunhua Shen, Yuning Jiang, and Lei Li.
\newblock Solo: Segmenting objects by locations.
\newblock In \emph{ECCV}, 2020{\natexlab{a}}.

\bibitem[Wang et~al.(2020{\natexlab{b}})Wang, Zou, Yi, Bailey, Ma, and
  Gu]{MART}
Yisen Wang, Difan Zou, Jinfeng Yi, James Bailey, Xingjun Ma, and Quanquan Gu.
\newblock Improving adversarial robustness requires revisiting misclassified
  examples.
\newblock In \emph{ICLR}, 2020{\natexlab{b}}.

\bibitem[Wang et~al.(2023)Wang, Pang, Du, Lin, Liu, and Yan]{wang2023better}
Zekai Wang, Tianyu Pang, Chao Du, Min Lin, Weiwei Liu, and Shuicheng Yan.
\newblock Better diffusion models further improve adversarial training.
\newblock In \emph{ICML}, 2023.

\bibitem[Wei et~al.(2023)Wei, Wang, Guo, and Wang]{CFA}
Zeming Wei, Yifei Wang, Yiwen Guo, and Yisen Wang.
\newblock Cfa: Class-wise calibrated fair adversarial training.
\newblock In \emph{CVPR}, 2023.

\bibitem[Wen et~al.(2021)Wen, Yiu, and Hui]{wen2021defending}
Jing Wen, Siu-Ming Yiu, and Lucas~CK Hui.
\newblock Defending against model inversion attack by adversarial examples.
\newblock In \emph{IEEE CSR}, pages 551--556, 2021.

\bibitem[Wu et~al.(2020)Wu, Xia, and Wang]{AWP}
Dongxian Wu, Shu-Tao Xia, and Yisen Wang.
\newblock Adversarial weight perturbation helps robust generalization.
\newblock In \emph{NeurIPS}, 2020.

\bibitem[Xie et~al.(2020)Xie, Tan, Gong, Wang, Yuille, and
  Le]{xie2020adversarial}
Cihang Xie, Mingxing Tan, Boqing Gong, Jiang Wang, Alan~L Yuille, and Quoc~V
  Le.
\newblock Adversarial examples improve image recognition.
\newblock In \emph{CVPR}, 2020.

\bibitem[Yin et~al.(2023)Yin, Yao, Shi, Du, and Xiao]{Robust_Gap_Paper}
Shenglin Yin, Kelu Yao, Sheng Shi, Yangzhou Du, and Zhen Xiao.
\newblock Again: Adversarial training with attribution span enlargement and
  hybrid feature fusion.
\newblock In \emph{CVPR}, 2023.

\bibitem[Zagoruyko and Komodakis(2016)]{WRN_NET}
Sergey Zagoruyko and Nikos Komodakis.
\newblock Wide residual networks.
\newblock \emph{arXiv preprint arXiv:1605.07146}, 2016.

\bibitem[Zeng et~al.(2024)Zeng, Li, Liu, Gao, Jiang, Tang, Hu, Liu, and
  Zhang]{zeng2024controllable}
Bohan Zeng, Shanglin Li, Xuhui Liu, Sicheng Gao, Xiaolong Jiang, Xu Tang, Yao
  Hu, Jianzhuang Liu, and Baochang Zhang.
\newblock Controllable mind visual diffusion model.
\newblock In \emph{AAAI}, 2024.

\bibitem[Zhang et~al.(2019)Zhang, Yu, Jiao, Xing, Ghaoui, and Jordan]{TRADES}
Hongyang Zhang, Yaodong Yu, Jiantao Jiao, Eric Xing, Laurent~El Ghaoui, and
  Michael Jordan.
\newblock Theoretically principled trade-off between robustness and accuracy.
\newblock In \emph{ICML}, 2019.

\bibitem[Zhang et~al.(2024)Zhang, Zhao, Chen, Jiang, and Chen]{zhangfeature}
Tianren Zhang, Chujie Zhao, Guanyu Chen, Yizhou Jiang, and Feng Chen.
\newblock Feature contamination: Neural networks learn uncorrelated features
  and fail to generalize.
\newblock In \emph{ICML}, 2024.

\end{thebibliography}
